\documentclass[aps,amsmath,pre,floatfix,superscriptaddress]{revtex4-2}
\usepackage{graphicx}
\usepackage{geometry}
\usepackage{wrapfig}
\usepackage{comment}
\usepackage{amsmath}
\usepackage[export]{adjustbox}
\usepackage{float}
\usepackage{amssymb}
\usepackage[utf8]{inputenc}
\usepackage{afterpage}
\usepackage{lipsum}  
\usepackage{appendix}
\usepackage{natbib}
\usepackage{subcaption}
\usepackage{amsmath}
\usepackage{setspace}

\usepackage{hyperref}
\hypersetup{colorlinks=true, linktoc=all, linkcolor=blue, linktocpage, citecolor=blue}

\usepackage{geometry}\usepackage{geometry}

\usepackage[normalem]{ulem}
\graphicspath{{figs/}}

\begin{document}
\title{Trust  AI Regulation? Discerning users are vital to build trust and effective AI regulation}
\author{Zainab Alalawi$^{1}$}
\author{Paolo Bova$^{1}$}
\author{Theodor Cimpeanu$^{2}$}
\author{Alessandro Di Stefano$^{1}$}
\author{Manh Hong Duong$^{3}$}
\author{Elias Fernández Domingos$^{4,5}$}
\author{The Anh Han$^{1,*}$}
\author{Marcus Krellner$^{2}$}
\author{Bianca Ogbo$^{1}$}
\author{Simon T. Powers$^{6}$}
\author{Filippo Zimmaro$^{7,8}$}



\maketitle
	{\footnotesize
		\noindent
		$^{1}$ School Computing, Engineering and Digital Technologies, Teesside University\\
		$^{2}$  School of Mathematics and Statistics, University of St Andrews\\
        $^{3}$ School of Mathematics, University of Birmingham\\
        $^{4}$ Machine Learning Group, Universit\'e libre de Bruxelles\\ 
        $^{5}$  AI Lab, Vrije Universiteit Brussel\\
	    $^{6}$ School of Computing, Engineering and the Built Environment, Edinburgh Napier University\\ 
        $^{7}$ Department of Mathematics, University of Bologna\\
        $^{8}$ Department of Computer Science, University of Pisa\\
  $^\star$ Corresponding author: The Anh Han (T.Han@tees.ac.uk)
	}

\section*{abstract}
There is general agreement that some form of regulation is necessary both for AI creators to be incentivised to develop trustworthy systems, and for users to actually trust those systems. But there is much debate about what form these regulations should take and how they should be implemented. Most work in this area has been qualitative, and has not been able to make formal predictions. Here, we propose that evolutionary game theory can be used to quantitatively model the dilemmas faced by users, AI creators, and regulators, and provide insights into the possible effects of different regulatory regimes. We show that creating trustworthy AI and user trust requires regulators to be incentivised to regulate effectively.
We demonstrate the effectiveness of two mechanisms that can achieve this. The first is where governments can recognise and reward regulators that do a good job. In that case, if the AI system is not too risky for users then some level of trustworthy development and user trust evolves. We then consider an alternative solution, where users can condition their trust decision on the effectiveness of the regulators. This leads to effective regulation, and consequently the development of trustworthy AI and user trust, provided that the cost of implementing regulations is not too high. Our findings highlight the importance of considering the effect of different regulatory regimes from an evolutionary game theoretic perspective.


\newpage

\begin{figure*}
\begin{center}
\includegraphics[width=1\textwidth]{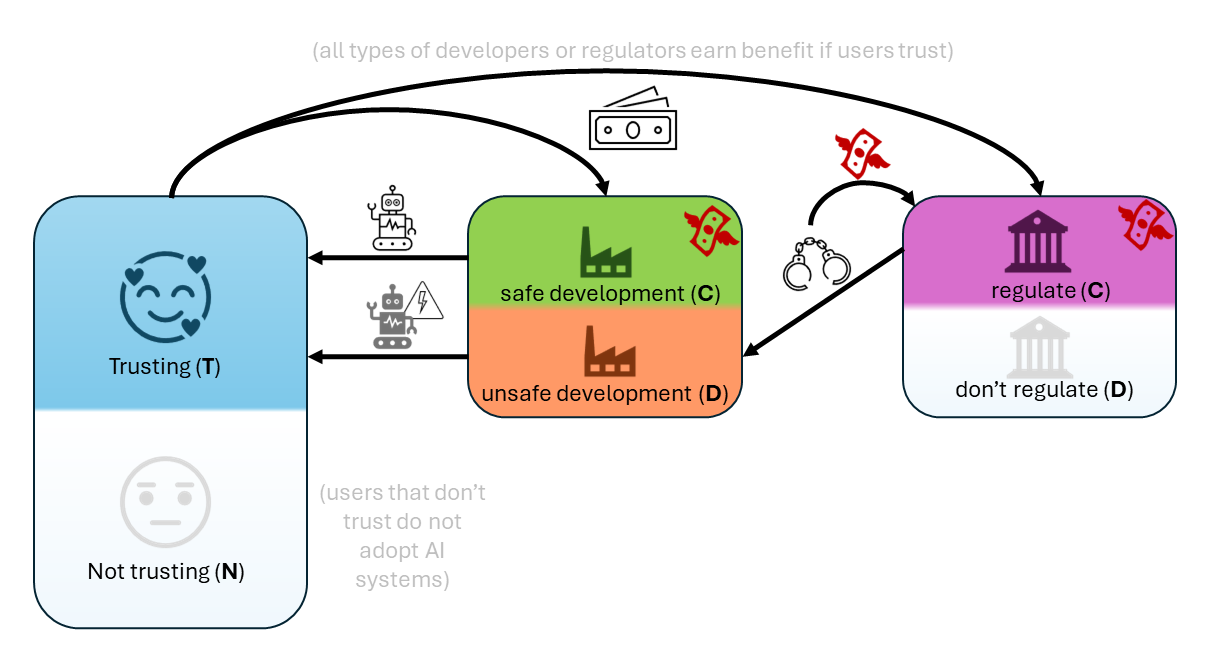}
\caption{\textbf{Core features}. The figure schematically illustrates the core features of the three-population model of AI governance. Users can either trust (T) or not trust (N) the AI system, in which case they do not adopt the system and get zero benefit. Creators can either defect by creating unsafe AI products (D) or cooperate by creating safe ones (C), which entails additional costs. Applying regulations (C) also comes with some costs, while punishing defecting creators requires further costs.}
\label{fig1}
\end{center}
\end{figure*}

\section{Introduction}

Debates are taking place across the world about what kind of regulation should apply to the development of artificial intelligence (AI) systems. Governments want \emph{trustworthy} AI systems to be developed, and for users to actually \emph{trust} these systems -- this is exemplified by the aims of the EU AI act \cite{laux2024trustworthy}. Regulation is typically assumed to be the way to achieve this \cite{powers2023stuff}. Economic history supports this assumption; safety critical systems such as automotive vehicles and medical devices are indeed regulated in this way. But it is less clear what forms the regulations should take and who should implement them. See, for example, proposals in the EU compared to the USA on the amount of restrictions that should be placed on AI creators \cite{siegmann2022brussels, baker2023executive}. Moreover, there is the key question of \emph{who} should create and enforce the regulations? Should the regulations be \emph{public} or \emph{private} \cite{tallberg2023global}? That is, should they be created by government bodies, or by industrial organisations and accrediting bodies such as the IEEE? And once they have been created, who will enforce them? Enforcement could involve actions such as auditing source code, training runs, training data, and monitoring for consumer complaints. This could be done by government sector organisations, but it could also be done by third-party auditors \cite{clark2019regulatory}. 

To help governments choose the most suitable kind of regulatory framework, we need to be able to predict the effects of different regulatory systems. Most of the discourse around this at the moment is qualitative and does not lead to formal predictions \cite{dafoe2023ai, anderljung2023frontier, hadfield2023regulatory, alaga2023coordinated}. This limits the ability of governments, technology companies, and citizens to foresee what the effects of different regulatory systems might be \cite{pitt2023chatsh}. There is a small amount of literature on AI race modelling \cite{armstrong2016racing,lacroix2022tragedy, jensen2023industrial1}, including from the perspective of (evolutionary) game theory \cite{han2020regulate, han2022institutional, cimpeanu2022artificial, bova2023tale}, but this has not considered how different regulatory mechanisms influence both user trust and the compliance of companies with AI safety considerations. To address this, we propose that evolutionary game theory \cite{hofbauer1998evolutionary} can be used to formally model the effects of different regulatory systems in terms of their incentives on tech companies, end users, and regulators.

We present a framework for formalising the strategic interaction between users, AI system creators, and regulators as a game (Fig.~1). This game captures three key decisions that these actors face:
\begin{enumerate}
\item \textbf{Users}: do they trust and hence use an AI system or not?
\item \textbf{AI creators}: do they follow a safety optimal development path in compliance with regulations, or do they pursue a competitive development path that violates regulations in a race to the market?
\item \textbf{Regulators}: do they invest in monitoring AI system creators and enforcing the regulations effectively, or do they cut corners to save the costs of doing this? 
\end{enumerate}

This highlights the dilemmas facing users, creators, and regulators. Users can benefit from using an AI system, but also run the risk that the system may not act in their best interest, i.e. may not be trustworthy \cite{han2021or}. This follows from the fact that AI creators are themselves in competition with each other, as highlighted by the current ``AI race'' to develop artificial general intelligence \cite{armstrong2016racing, askell2019role, anderljung2023frontier, cottier2024who, grant2024how}. Consequently, we cannot assume that creators will always act in the best interests of their users by complying with regulations and developing systems worthy of user trust. Finally, regulators themselves may be self-interested. This may occur when governments delegate the enforcement of regulations to other actors, such as private audit firms \cite{cihon2021ai,clark2019regulatory, hadfield2023regulatory}. This kind of delegation may reduce government costs, but it also introduces a principal agent problem \cite{north1990institutions}: regulators are themselves agents with their own profit maximising goals.

To analyse the model, we use the methods of evolutionary game theory. Evolutionary game theory is based on the idea that agents can learn behaviours that benefit them from social learning, i.e. by copying the behaviour of other agents in their population that are doing better than themselves. This avoids the need to assume that the agents are fully rational and have complete information. In our model, we consider three populations corresponding to the three actors: users, creators, and regulators. 

Our analysis demonstrates how incentives for regulators are important. Governments desire that all creators produce trustworthy AI systems, and all users trust these systems. Such a state cannot be reached if regulators that do their job properly cannot be distinguished from regulators that cut corners. This holds regardless of the severity of punishment for defecting creators.  

We consider two possible institutional solutions to this problem. First, we show that if governments can provide rewards to regulators that do a good job, and use of the AI system is not too risky for users, then some level of trustworthy development and trust by users occurs. We then consider an alternative solution, where users may condition their trust decision on the effectiveness of the regulators, for example, where information about the past performance of the regulators is available. This leads to effective regulation, and consequently the development of trustworthy AI and user trust, provided that the cost of implementing regulations is not too high.



    
\section{Models and Methods}

\subsection{Three population model of AI governance}

We create a \textbf{baseline model} of an AI development ecosystem \cite{powers2023stuff}. It is similar to current models of regulations for information systems, e.g. the General Data Protection Regulations (GDPR)  and the AI Act of the European Union. 
The model involves three populations representing the three actors in the ecosystem: AI users, AI system creators, and regulators. In each population, individuals can choose different options (also called strategies). A user can decide to trust (T) or not (N) an AI system: This is a combination of trust in the regulator and the creator. The creator either complies (C) with the rules set out by the regulators or not (D). The regulator chooses whether to create rules and enforce compliance (C) or not (D). 

The individual payoff earned in any one encounter (also called a game) depends on the strategy of the participating individuals. In each game, one user, one developer, and one regulator participate. If the user trusts and adopts an AI system when both the developer and the regulator cooperate, the user benefits significantly from AI adoption, denoted by $b_U$. However, if AI adoption occurs when the developer defects, the user is affected by unsafe AI, gaining a reduced or even negative benefit, denoted by $\epsilon \times b_U$, where $\epsilon \in [-\infty, 1]$. This parameter, $\epsilon$, also represents a \textit{risk factor} that users take when adopting the AI system. 
For the system creator, since it takes more time and effort to comply with the precautionary requirements, we assume that playing C requires an extra cost $c_P$, compared to playing D (the cost for this is normalised to 0). 
Regarding the regulators, they earn the benefit $b_R$ when the user trusts and adopts AI. Since it is costly to create rules and development technologies to capture unsafe development,  we assume that playing C requires a extra cost $c_R$, compared to playing D (the cost for this is normalised to 0). 
The model strategies and parameters are summarised in Table \ref{tbl:PModel}.

\begin{table}[h!]
    \centering
    \normalsize
    \caption{AI Governance model (User \textit{Us.}, Creator \textit{Cr} and Regulator \textit{Re}).}
    \label{tbl:PModel}

    \begin{tabular}{|p{0.7\textwidth}|p{0.09\textwidth}|}
        \hline
        \textbf{Parameters\textquotesingle description} &  \textbf{Symbol} \\
        \hline \hline
            Users trust (T) or not (N) trust an AI system – this is a combination of trust in regulators and trust in creators & \textit{T}, \textit{N} \\
        \hline
            Creators comply with the rules (C) or not (D) as set out by the Regulators & \textit{C}, \textit{D} \\
        \hline
            Regulators create rules and ensure compliance (C) or not (D) & \textit{C}, \textit{D} \\
        \hline
            Benefit users get from trust and adopt the AI system & $b_U$ \\
        \hline
             Fraction of user benefit when creators play \textit{D}, where $\varepsilon$ in [-$\infty$,1], also referred to as the (inverse) risk factor users take when adopting an AI & $\varepsilon$ \\
       \hline
             Benefit the creator gets from selling the product & $b_P$\\
       
        \hline
            Additional cost of creating safe AI (the cost of creating unsafe AI is normalised to 0) & $c_P$\\
        \hline
          Funding for regulators (which is only  generated upon  AI adoption) & $b_R$\\
        \hline
            The cost of developing rules and unsafe capture technologies (the cost of not doing this is normalised to 0) & $c_R$\\
        \hline
            The cost of institutional punishment & $v$\\
        \hline
            The impact of institutional punishment & $u$\\
        \hline
        \end{tabular}
    
     \bigskip{\vspace{-0.1cm}}
    
    \begin{tabular}
        {|p{0.045\textwidth}|p{0.05\textwidth}|p{0.05\textwidth}|| p{0.2\textwidth} | p{0.2\textwidth}| p{0.2\textwidth}|}
       \hline
       \multicolumn{3}{|c||}{\textbf{Strategies}} &  \multicolumn{3}{ c |}{\textbf{Payoffs}} \\
          \hline
        \textbf{Us.} & \textbf{Cr} & \textbf{Re} & \textbf{User} & \textbf{Creator} & \textbf{Regulator} \\
        \hline \hline
         \textit{T} & \textit{C} & \textit{C} & $b_U$ & $b_P-c_P$ & $b_R- c_R$ \\
        \hline
         \textit{T} & \textit{C} & \textit{D} & $b_U$ & $b_P - c_P$ & $b_R$\\
        \hline
         \textit{T} & \textit{D} & \textit{C} & $\varepsilon b_U$ & $b_P-u$ & $b_R-c_R-v$\\
        \hline
         \textit{T} & \textit{D} & \textit{D} & $\varepsilon b_U$ & $b_P$ & $b_R$ \\
        \hline
         \textit{N} & \textit{C} &\textit{C} & 0 & $-c_P$ & $-c_R$\\
       \hline
         \textit{N} & \textit{C} & \textit{D} & 0 & $- c_P$ & 0 \\
       \hline
        \textit{N} & \textit{D} &\textit{C} & 0 & 0 & -$c_R$\\
       \hline
         \textit{N} & \textit{D} & \textit{D} & 0 &  0 &  0\\
      \hline
     \end{tabular}

\end{table}

As shown in our analysis below, in this system, regulators are always better off not cooperating, and trust is rarely encouraged. 
Thus, below we consider two extended models that might enable investment of high quality regulation, safe development, and users' trust.  
\begin{enumerate}
    \item First, we consider that the cooperative regulator is rewarded an amount of $b_{fo}$ if they catch defective creators (when users trust and adopt AI). The payoff matrix is summarised in Table \ref{tbl:PModel2}. 
    \item Second, we assume that regulators' reputation is publicly available and users can act conditionally on whether the regulators' reputation is good or not; see Table \ref{tbl:PModel3}.  Namely, the CT strategy means users do not trust when the regulator's reputation is bad.

\end{enumerate}

In the general case, this leads to a frequency-dependent dynamics in each of the three populations, where the dynamical process of strategy adoption assumes that those individuals that are more fit are more often imitated by their peers, an adaptive scheme akin to social learning (see Methods below).

\begin{figure*}
\begin{center}
\includegraphics[width=1\textwidth]{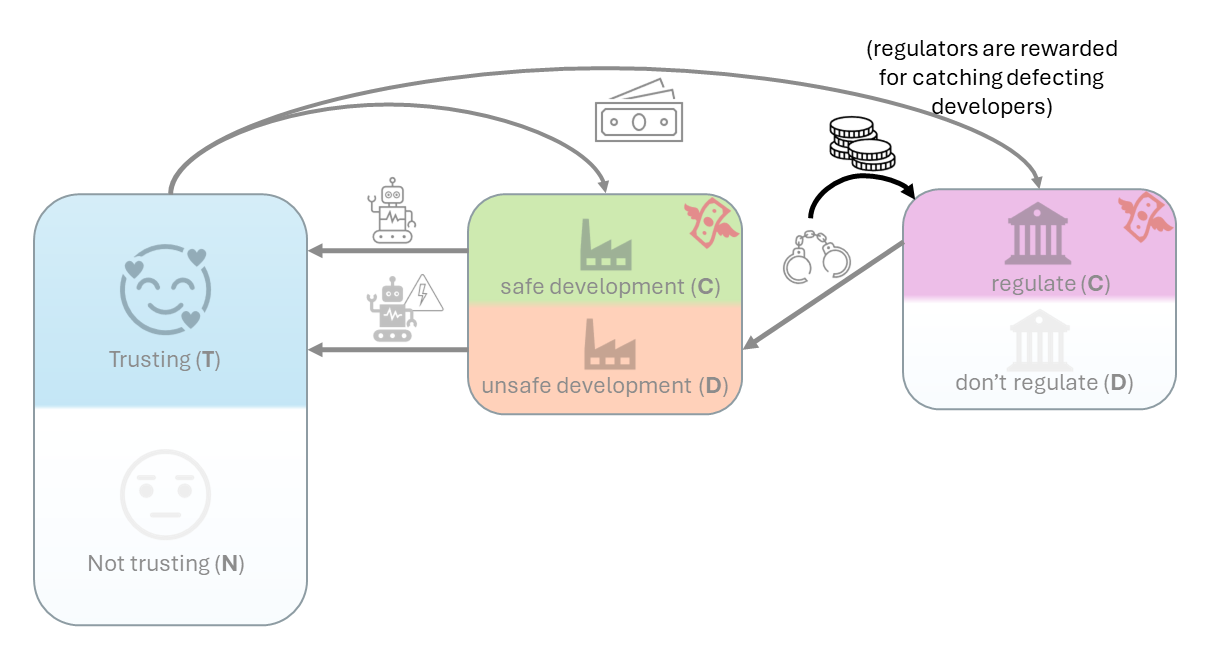}
\caption{\textbf{Reward for regulators}. The figure shows the changes to the original model. Regulators gain extra reward for capturing unsafe development.}
\label{fig2}
\end{center}
\end{figure*}

\begin{table}[h!]
    \centering
    \normalsize
    \caption{AI Governance extended model (User \textit{Us.}, Creator \textit{Cr} and Regulator \textit{Re}), where regulators are rewarded for capturing unsafe creators (when users trust and adopt).}
    \label{tbl:PModel2}
        \begin{tabular}
        {|p{0.045\textwidth}|p{0.05\textwidth}|p{0.05\textwidth}|| p{0.2\textwidth} | p{0.2\textwidth}| p{0.2\textwidth}|}
       \hline
       \multicolumn{3}{|c||}{\textbf{Strategies}} &  \multicolumn{3}{ c |}{\textbf{Payoffs}} \\
          \hline
        \textbf{Us.} & \textbf{Cr} & \textbf{Re} & \textbf{User} & \textbf{Creator} & \textbf{Regulator} \\
        \hline \hline
         \textit{T} & \textit{C} & \textit{C} & $b_U$ & $b_P-c_P$ & $b_R- c_R$ \\
        \hline
         \textit{T} & \textit{C} & \textit{D} & $b_U$ & $b_P - c_P$ & $b_R$\\
        \hline
         \textit{T} & \textit{D} & \textit{C} & $\varepsilon b_U$ & $b_P-u$ & $b_R-c_R-v {+b_{fo}}$\\
        \hline
         \textit{T} & \textit{D} & \textit{D} & $\varepsilon b_U$ & $b_P$ & $b_R$ \\
        \hline
         \textit{N} & \textit{C} &\textit{C} & 0 & $-c_P$ & $-c_R$\\
       \hline
         \textit{N} & \textit{C} & \textit{D} & 0 & $- c_P$ & 0 \\
       \hline
        \textit{N} & \textit{D} &\textit{C} & 0 & 0 & -$c_R$\\
       \hline
         \textit{N} & \textit{D} & \textit{D} & 0 &  0 &  0\\
      \hline
     \end{tabular}

\end{table}

\begin{figure*}
\begin{center}
\includegraphics[width=1\textwidth]{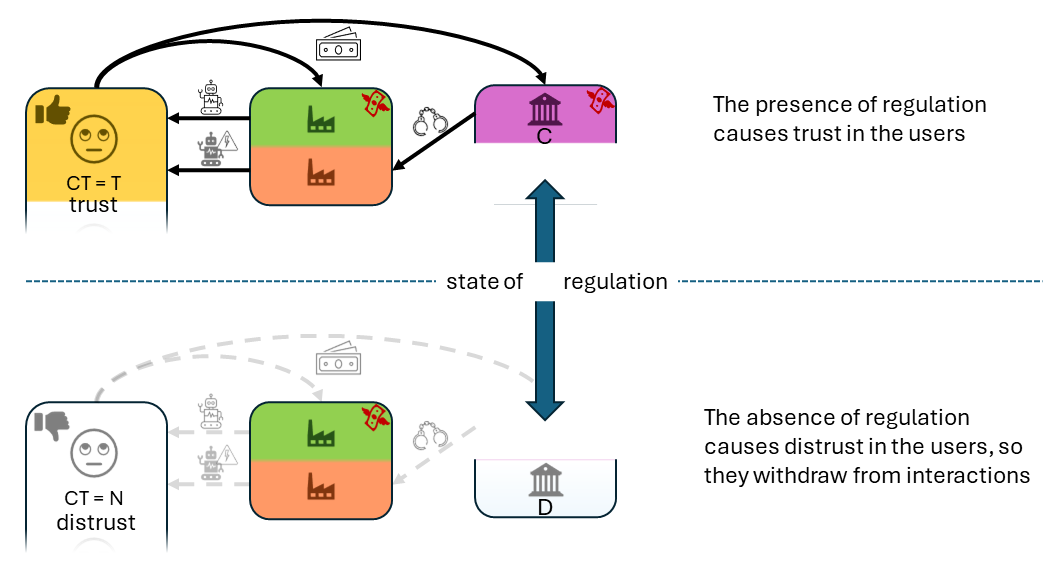}
\caption{\textbf{Conditional trust}. The figure shows how the state of regulation changes the behaviour of users, which impacts the whole system.}
\label{fig3}
\end{center}
\end{figure*}

\begin{table}[h!]
    \centering
    \normalsize
    \caption{AI Governance extended model---conditional trust (User \textit{Us.}, Creator \textit{Cr} and Regulator \textit{Re}). In this model, regulators' reputation are assumed to be publicly available before the game, and users can act conditionally on it: CT does not trust if regulator has a bad reputation. }
    \label{tbl:PModel3}
        \begin{tabular}
        {|p{0.045\textwidth}|p{0.045\textwidth}|p{0.045\textwidth}|| p{0.2\textwidth} | p{0.2\textwidth}| p{0.2\textwidth}|}
       \hline
       \multicolumn{3}{|c||}{\textbf{Strategies}} &  \multicolumn{3}{ c |}{\textbf{Payoffs}} \\
          \hline
        \textbf{Us.} & \textbf{Cr} & \textbf{Re} & \textbf{User} & \textbf{Creator} & \textbf{Regulator} \\
        \hline \hline
         \textit{CT} & \textit{C} & \textit{C} & $b_U$ & $b_P-c_P$ & $b_R- c_R$ \\
        \hline
         \textit{CT} & \textit{C} & \textit{D} &  {$0$} &  {$- c_P$} &  {$0$}\\
        \hline
         \textit{CT} & \textit{D} & \textit{C} & $\varepsilon b_U$ & $b_P-u$ & $b_R-c_R-v+b_{fo}$\\
        \hline
         \textit{CT} & \textit{D} & \textit{D} &  {$0$} &  {$0$} &  {$0$} \\
        \hline
         \textit{N} & \textit{C} &\textit{C} & 0 & $-c_P$ & $-c_R$\\
       \hline
         \textit{N} & \textit{C} & \textit{D} & 0 & $- c_P$ & 0 \\
       \hline
        \textit{N} & \textit{D} &\textit{C} & 0 & 0 & -$c_R$\\
       \hline
         \textit{N} & \textit{D} & \textit{D} & 0 &  0 &  0\\
      \hline
     \end{tabular}

\end{table}

\newpage
\subsection{Methods}
\subsubsection{Stochastic dynamics for finite populations}
\textbf{The baseline model}.
We consider three different well-mixed populations of Users (U), Creators (C) and Regulators (R) of sizes, respectively $N_U$, $N_C$ and $N_R$. 
Let $x$ be the fraction of users that trust the AI system. Let $y$ and $z$ be respectively the fraction of Creators and Regulators that cooperate. Each game involves an  individual  randomly  drawn  from  each  population. The fitness that a user, regulator and creator obtains in each game is respectively given by
\begin{align}
f^U_{X\in\{T,N\}}&=yz P^U_{XCC}+(1-y)z P^{U}_{XDC}+y(1-z) P^U_{XCD}+(1-y)(1-z) P^U_{XDD}, \label{eq: UTN}  \\
  f^C_{Y\in\{C,D\}}&=xz P^C_{TYC}+(1-x)z P^C_{NYC}+x(1-z) P^C_{TYD}+(1-x)(1-z)P^C_{NYD}, \label{eq: CCD}\\
  f^R_{Z\in\{C,D\}}&=xy P^R_{TCZ}+(1-x)z P^R_{NCZ}+x(1-y) P^R_{TDZ}+(1-x)(1-y)P^R_{NDZ}.\label{eq: RCD}
\end{align}

We now derive the fitness/average payoffs for each model. 
First, for  the baseline model, from \eqref{eq: UTN} and the payoff table  \ref{tbl:PModel}, we calculate explicitly the difference of the fitness between two strategies in users:
\begin{align}
f^U_T-f^U_N&= yz(P^U_{TCC}-P^U_{NCC})+(1-y)z (P^U_{TDC}-P^U_{TDC})+y(1-z) (P^U_{TCD}-P^U_{NCD})+(1-y)(1-z) (P^U_{TDD}-P^U_{NDD})\notag
\\&=yz b_U+(1-y)z(\varepsilon b_U)+y(1-z)b_U+(1-y)(1-z)(\varepsilon b_U)\notag
\\&=b_U\Big(yz+\varepsilon(1-y)z+y(1-z)+\varepsilon(1-y)(1-z)\Big)\notag
\\&=b_U(y+\varepsilon(1-y)).\label{eq: differnce T and N}
\end{align}
Similarly, the difference of the fitness between two strategies in creators is
\begin{align}
f^C_C-f^C_{D}&=  xz (P^C_{TCC}-P^C_{TDC})+(1-x)z (P^C_{NCC}-P^C_{NDC})+x(1-z) (P^C_{TCD}-P^C_{TDD})+(1-x)(1-z)(P^C_{NCD}-P^C_{NDD})\notag
\\&=xz \Big(b_P-c_P-(b_P-u)\Big)+(1-x)z (-c_P)+x(1-z) (b_P-c_P-b_P)+(1-x)(1-z)(-c_P)\notag
\\&=xz (u-c_P)+(1-x)z (-c_P)+x(1-z) (-c_P)+(1-x)(1-z)(-c_P)\notag
\\&=-c_P+u x z. \label{eq: difference C and D for creators}
\end{align}
Finally, the difference of the fitness between two strategies in regulators is
\begin{align}
f^R_{C}-f^R_{D}&= xy (P^R_{TCC}-P^R_{TCD})+(1-x)y (P^R_{NCC}-P^R_{NCD})+x(1-y) (P^R_{TDC}-P^R_{TDD})+(1-x)(1-y)(P^R_{NDC}-P^R_{NDD})  \notag
\\&=xy\Big((b_R-c_R)-b_R\Big)+(1-x)y(-c_R)+x(1-y)\Big(b_R-c_R-v-b_R\Big)+(1-x)(1-y)(-c_R)\notag
\\&=xy(-c_R)+(1-x)y(-c_R)+x(1-y)(-c_R-v)+(1-x)(1-y)(-c_R)\notag
\\&=-c_R-x(1-y)v.\label{eq: difference C and D for regulators}
\end{align}
We notice that $f^R_{C}-f^R_{D}$ is always negative. Thus, it is always better off for the regulator to defect, that is, to provide regulator rules.

Now, for the extended model where the reward $b_{fo}$ is provided to regulators for capturing unsafe creators and if they catch defective creators (when users adopt only). 
Then only the payoff difference $f^R_{C}-f^R_{D}$ is changed to
\begin{equation}
\label{eq: difference payoff R extended}
f^R_{C}-f^R_{D}=-c_R+x(1-y)(b_{fo}-v).    
\end{equation}

Finally, when users can make a conditional decision based on the regulator's reputation, we have 
\begin{equation}
\begin{split}
    &f^U_T-f^U_N = b_Uz(y+\epsilon (1-y))\\ 
    &f^C_C-f^C_{D} = - c_P + uxz \\ 
    &f^R_C-f^R_{D} = -c_R + b_Rx + (b_{fo}-v)x(1-y) 
\end{split}
\end{equation}

For a finite population setting, at each time step, a randomly selected individual A, with fitness $f_A$, may adopt a different strategy by imitating a randomly chosen individual B from the same population (with fitness $f_B$) with probability given by the Fermi distribution
\[
p=[1+e^{-\beta(f_B-f_A)}]^{-1},
\]
where $\beta\geq 0$ is the strength of selection. $\beta = 0$ corresponds to neutral drift where imitation decisions are random, while for large $\beta \rightarrow \infty$, the imitation decision becomes increasingly deterministic.

In the absence of mutations or exploration, the end states of evolution are inevitably monomorphic: once such a state is reached, it cannot be escaped through imitation. We thus further assume that with a certain mutation probability,  an agent switches randomly to a different strategy without imitating another agent.  In the limit of small mutation rates, the dynamics will proceed with, at most, two strategies in the population, such that the behavioural dynamics can be conveniently described by a Markov chain, where each state represents a monomorphic population, whereas the transition probabilities are given by the fixation probability of a single mutant \citep{key:imhof2005,key:novaknature2004,domingos2023egttools}. The resulting Markov chain has a stationary distribution, which characterises the average time the population spends in each of these monomorphic end states.


Now, the probability to change the number $k$ of agents using strategy A by $\pm$ one in each time step can be written as ($Z$ is the population size) \citep{traulsen2006} 
\begin{equation} 
T^{\pm}(k) = \frac{Z-k}{Z} \frac{k}{Z} \left[1 + e^{\mp\beta[f_A(k) - f_B(k)]}\right]^{-1}.
\end{equation}
The fixation probability of a single mutant with a strategy A in a population of $(Z-1)$ agents using B is given by \citep{traulsen2006,key:novaknature2004}
\begin{equation} 
\label{eq:fixprob} 
\rho_{B,A} = \left(1 + \sum_{i = 1}^{Z-1} \prod_{j = 1}^i \frac{T^-(j)}{T^+(j)}\right)^{-1}.
\end{equation} 

The transition matrix $\Lambda$ corresponding to the set of $\left\{1,\ldots ,s\right\}$ strategies is given by:
\begin{eqnarray}\label{eq:2.6}
\Lambda_{ij,j\neq i}=\frac{\rho_{ji}}{3}\hspace{1mm} \text{ and } \hspace{1mm} \Lambda_{ii}= 1- \sum_{j=1,j\neq i}^s \Lambda_{ij}.
\end{eqnarray}
  Fixation probability $\rho_{ij}$ denotes the likelihood that a population transitions from a state $i$ to a different state $j$ when a mutant of one of the populations adopts an alternate strategy \textit{s}. The fixation probability is divided by the number of populations (3) representing the interaction of three players at a time \cite{encarnaccao2016paradigm,alalawi2019pathways}. 

\subsubsection{Population dynamics for infinite populations: The multi-population replicator dynamics}
In this section, we recall the framework of the replicator dynamics for multi-populations \cite{taylor1979evolutionarily,bauer2019stabilization}. To describe the dynamics, we consider a set of $m$ different populations ($m$ is some positive integer), which are infinitely large and well-mixed. Each population $i$, $i=1,\ldots m$, consists of $n_i$ ($n_i$ is some positive integer) different strategies (types). Let $x_{ij}, 1\leq i\leq m, 1\leq j\leq n_i$, be the frequency of the strategy $j$ in the population $i$. We denote by $x_i=(x_{ij})_{j=1}^{n_i}$, which is the collection of all strategies in the population $i$, and $x=(x_1,\ldots, x_m)$, which is the collection of all strategies in all populations. 

For each $i\in\{1,\ldots, m\}$ and $j\in\{1,\ldots, n_i\}$, let $f_{ij}(x)$ be the fitness (reproductive rate) of the strategy $j$ in the population $i$. This fitness is obtained when the strategy $j$ interacts with all other strategies in all populations; thus, it depends on all the strategies in the populations. The average fitness of the population $i$ is defined by
\[
\bar{f}_i(x)=\sum_{j=1}^{n_i} x_{ij} f_{ij}(x).
\]
The multi-population replicator dynamics is then given by
\begin{equation}
\label{eq: general replicator dynamics}
\dot{x}_{ij}=x_{ij} (f_{ij}(x)-\bar{f}_{i}(x)), \quad 1\leq i\leq m,\quad 1\leq j\leq n_i. 
\end{equation}
This is in general an ODE system of $\sum_{i=1}^m n_i$ equations. Noting, however that since $\sum_{j=1}^{n_i}x_{ij}=1$ for all $i=1,\ldots, m$, we can reduce the above system to a system of $\sum_{i=1}^m n_i-m$ equations. 

Now we  focus on the case when there are two strategies in each population (which is the case for our models of AI governance and trust in the present paper), that is $n_i=2$ for all $i=1,\ldots, n_i$. Let $\eta_i$ be the frequency of the first strategy in the population $i$, $i=1,\ldots, m$ (thus $1-\eta_i$ will be the frequency of the second strategy in the population $i$), let $\eta=(\eta_1,\ldots, \eta_m)$. Let $f_{1i}(\eta)$ and $f_{2i}(\eta)$ be the fitness of the first and second strategy in the population $i$. Since
\[
\bar{f}_i(\eta)=\eta_i f_{1i}(\eta)+(1-\eta_i) f_{2i}(\eta),
\]
we have
\[
f_{1i}(\eta)-\bar{f}_i(\eta)=f_{1i}(\eta)-(\eta_i f_{1i}(\eta)+(1-\eta_i) f_{2i}(\eta))=(1-\eta_i)(f_{1i}(\eta)-f_{2i}(\eta)).
\]
Thus we obtain the following system of equations
\begin{equation}
\label{eq: general replicator2}
\dot{\eta}_i=\eta_i(1-\eta_i)(f_{1i}(\eta)-f_{2i}(\eta)), \quad i=1,\ldots, m.
\end{equation}
This is a system of $m$ coupled nonlinear ordinary differential equations for $m$ variables.

In the subsequent sections, we employ \eqref{eq: general replicator2} to our models of AI governance trust, where the fitness are computed from the payoff matrix constructed in the models, see Tables~\ref{tbl:PModel}-\ref{tbl:PModel2}-\ref{tbl:PModel3}, where we assume that the payoffs are directly translated to the biological fitnesses (infinite strength of selection).

\section{Equilibrium analysis in infinite populations}

\subsubsection{The baseline model} 
We consider three different well-mixed populations of Users (U), Creators (C) and Regulators (R). Let $x$ be the frequency of users that trust the AI system. Let $y$ and $z$ be respectively the frequency of Creators and Regulators that cooperate.

The replicator dynamics is
\begin{subequations}
\label{eq: replicator dynamics baseline model}
\begin{align}
    \dot{x}&=x(1-x)(f^U_T-f^U_N)=x(1-x)b_U(y+\varepsilon(1-y)),\quad &x(0)=x_0\\ 
    \dot{y}&=y(1-y)(f^C_C-f^C_D)=y(1-y)(-c_P+u x z),\quad &y(0)=y_0\\ 
    \dot{z}&=z(1-z)(f^R_C-f^R_D)=z(1-z)(-c_R-x(1-y)v),\quad &z(0)=z_0
\end{align}
\end{subequations}
where $(x_0,y_0,z_0)\in [0,1]^3$ is the initial data. This is a complex, nonlinear system of three coupled ordinary differential equations (ODEs) for three variables. In general, the dynamical solutions $x(t), y(t), z(t)$ may show different behaviour depending on the values of the parameters and the initial data. We now focus on stationary solutions (equilibria), which describe the long-time behaviour of the dynamics.
\vspace{0.1in}

\paragraph{Stationary solutions (equilibrium points)}

Stationary solutions of \eqref{eq: replicator dynamics baseline model} are points $(x^*,y^*,z^*)\in [0,1]^3$ that make the right hand side of \eqref{eq: replicator dynamics baseline model} vanish.  It is clear that the vertices of the unit cube $(x,y,z)\in \{0,1\}^3$ are stationary solutions (we call them vertex equilibria). 

There may exist other, non-vertex, stationary solutions depending on the sign of $\varepsilon$ and the relation between $c_P$ and $u$.

If $\varepsilon> 0$, then $y+\varepsilon(1-y)>0$ for all $y\in[0,1]$. It implies that there are no other stationary solutions (except in the very specific case where $u=c_p$, in which $(x^*,y^*,z^*)\in \{0,1\}\times [0,1]\times \{0,1\}$ is a stationary solution). 

If $\varepsilon<0$, and $0<c_P\leq u$, then other, non-vertex, stationary solutions are given by
\[
y^*=\frac{\varepsilon}{\varepsilon-1}, \quad -c_P+ u x^* z^*=0, \quad z^*=1,
\]
which is
\begin{equation}
\label{eq: non-facet equil}
x^*=\frac{c_P}{u}, \quad y^*=\frac{\varepsilon}{\varepsilon-1}, \quad z^*=1.    
\end{equation}
\paragraph{Stability Analysis}
Next, we study the stability of the equilibria. To this end, we define
\begin{align*}
&F(x,y,z)=(F_1(x,y,z), F_2(x,y,z), F_3(x,y,z)), \quad F_1(x,y,z)=x(1-x)b_U(y+\varepsilon(1-y)), \\
& F_2(x,y,z)=y(1-y)(-c_P+u x z), \quad F_3(x,y,z)=z(1-z)(-c_R-x(1-y)v).
\end{align*}
Then \eqref{eq: replicator dynamics baseline model} can be reformulated as
\[
\dot{x}=F_1(x,y,z), \quad \dot{y}=F_2(x,y,z), \quad \dot{z}=F_3(x,y,z).
\]
The Jacobian matrix is 
\begin{align*}
DF(x,y,z)&=\begin{pmatrix}
\frac{\partial F_1}{\partial x}&\frac{\partial F_1}{\partial y}& \frac{\partial F_1}{\partial z}\\
\frac{\partial F_2}{\partial x}&\frac{\partial F_2}{\partial y}& \frac{\partial F_2}{\partial z}\\
\frac{\partial F_3}{\partial x}&\frac{\partial F_3}{\partial y}& \frac{\partial F_3}{\partial z}\\
\end{pmatrix}    
\\&=\begin{pmatrix}
b_U (1-2x)(y+\varepsilon(1-y))&b_U x(1-x)(1-\varepsilon)&0\\
u y(1-y) z& (1-2y)(-c_P+uxz)& u y(1-y) x\\
-z(1-z)(1-y)v& xv z(1-z)& (1-2z)(-c_R-x(1-y)v)
\end{pmatrix}.
\end{align*}
We recall that an equilibrium $(x^*,y^*,z^*)$ is asymptotically stable if all eigenvalues of $DF(x^*,y^*,z^*)$ have negative real parts.
\vspace{0.1in}

\paragraph{Stability Analysis for vertex equilibria}

We now determine the stability for vertex-equilibria, $(z^*,y^*,z^*)\in\{0,1\}^3$. For these equilibria, $DF(x^*,y^*,z^*)$ is always a diagonal matrix and is explicitly given by
\begin{equation*}
DF(x^*,y^*,z^*)=\begin{pmatrix}
b_U (1-2x^*)(y^*+\varepsilon(1-y^*))& 0 &0\\
0& (1-2y^*)(-c_P+ux^*z^*)& 0\\
0& 0& (1-2z^*)(-c_R-x^*(1-y^*)v)
\end{pmatrix}.
\end{equation*}
For instance at the origin $(x^*,y^*,z^*)=(0,0,0)$ we have
\[
DF(x,y,z)=\begin{pmatrix}
    \varepsilon b_U&0&0\\
    0&-c_P&0\\
    0&0&-c_R
\end{pmatrix}.
\]
For another example at $(x^*,y^*,z^*)=(1,0,0)$ then
\[
DF(x,y,z)=\begin{pmatrix}
    -\varepsilon b_U&0&0\\
    0&-c_P&0\\
    0&0&-c_R
\end{pmatrix}.
\]
Since for vertex equilibria, $DF(x^*,y^*,z^*)$ is diagonal and its eigenvalues read
\[
\lambda_1=b_U (1-2x^*)(y^*+\varepsilon(1-y^*)), \quad \lambda_2=(1-2y^*)(-c_P+ux^*z^*), \quad\lambda_3=(1-2z^*)(-c_R-x^*(1-y^*)v).
\]
Thus $(x^*,y^*,z^*)$ is stable only if
\[
\lambda_1<0, \quad \lambda_2<0, \quad \lambda_3<0.
\]
Since  $(-c_R-x^*(1-y^*)v)<0$, it follows that $\lambda_3<0$ iff $z^*=0$. For $z^*=0$, $\lambda_2=-c_P(1-2y^*)<0$ iff $y^*=0$. Then $\lambda_1=\varepsilon b_U(1-2x^*)$.\\
For $\varepsilon>0$, then $\lambda_1<0$ iff $x^*=1$. On the other hand, if $\varepsilon<0$, then $\lambda_1<0$ iff $x^*=0$.\\
In conclusion, for vertex equilibria, $(x^*,y^*,z^*)\in\{0,1\}^3$:
\begin{itemize}
    \item when $\varepsilon>0$, the equilibrium $(x^*, y^*,z^*)=(1,0,0)$ is stable, thus the users will fully trust (even if the creators and the regulators defect).
    \item when $\varepsilon<0$, the origin $(x^*,y^*,z^*)=(0,0,0)$ is stable, both regulators and creators defect and there is no trust from users.
\end{itemize}

\paragraph{Stability analysis for the non-vertex equilibrium}. For the non-vertex equilibrium \eqref{eq: non-facet equil}, the Jacobian matrix is of the form
\[
DF=\begin{pmatrix}
   0&b_U \frac{c_P}{u}(1-\frac{c_p}{u})(1-\varepsilon)&0\\
   -u\frac{\varepsilon}{(\varepsilon-1)^2}&0& -u\frac{\varepsilon}{(\varepsilon-1)^2}\frac{c_P}{u}\\
   0&0&c_R+\frac{c_P}{u}\frac{1}{1-\varepsilon} v
\end{pmatrix}.
\]
Thus it has one positive (recalling that $\varepsilon<0$ in this case) real root, which is the last diagonal entry. Therefore, the non-vertex equilibrium is not stable.

\subsubsection{Extended model: incentive for regulator to coperate}
\begin{figure}[h!]
    \centering
    \begin{subfigure}[b]{0.4\textwidth}
        \centering
        \includegraphics[width=\textwidth]{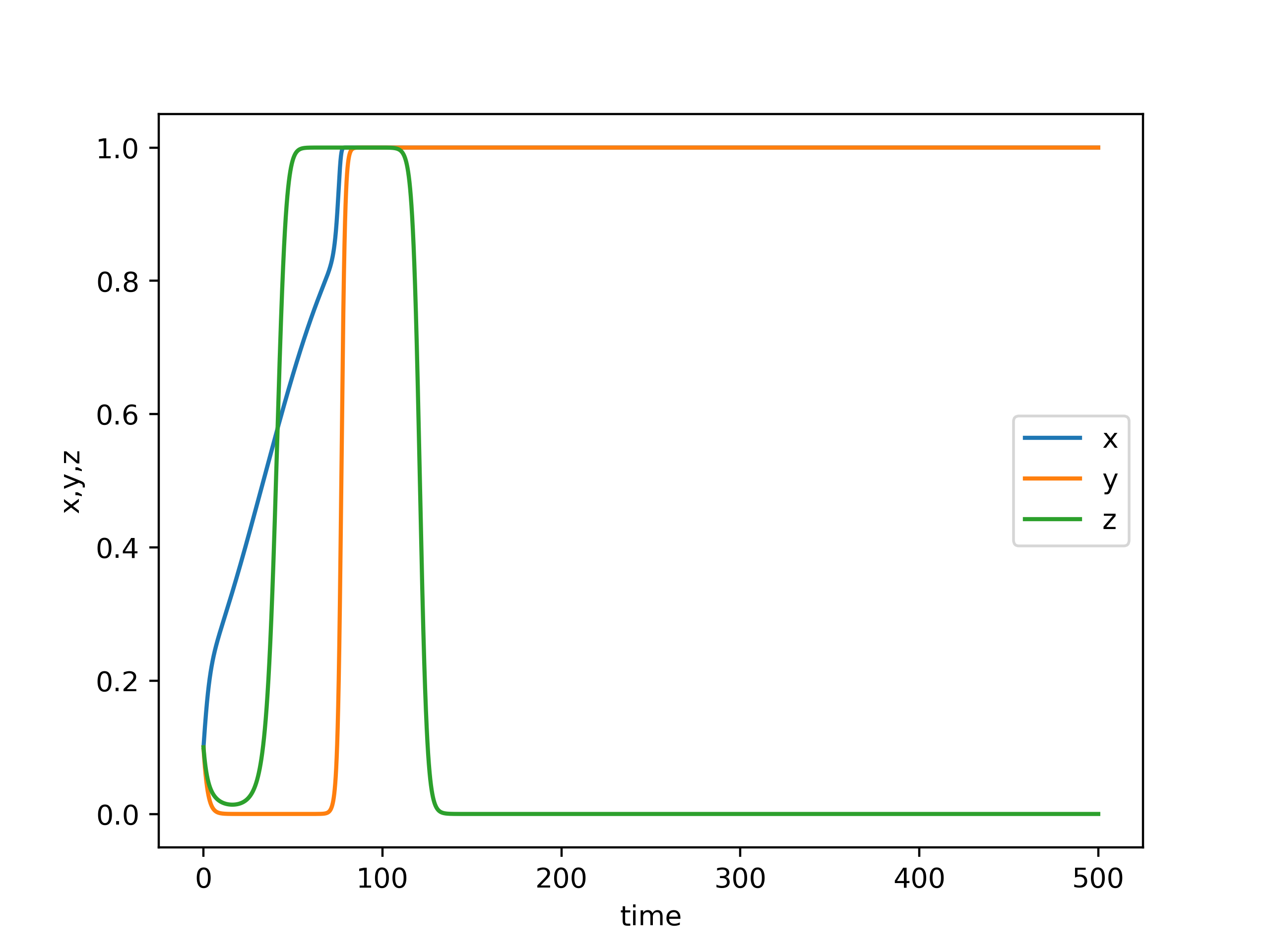}
        \caption{$x_0=y_0=z_0=0.1$, $\epsilon = 0.01$ }
        \label{fig:sub1a}
    \end{subfigure}
    \hspace{-0.3 cm}
    \begin{subfigure}[b]{0.4\textwidth}
        \centering
        \includegraphics[width=\textwidth]{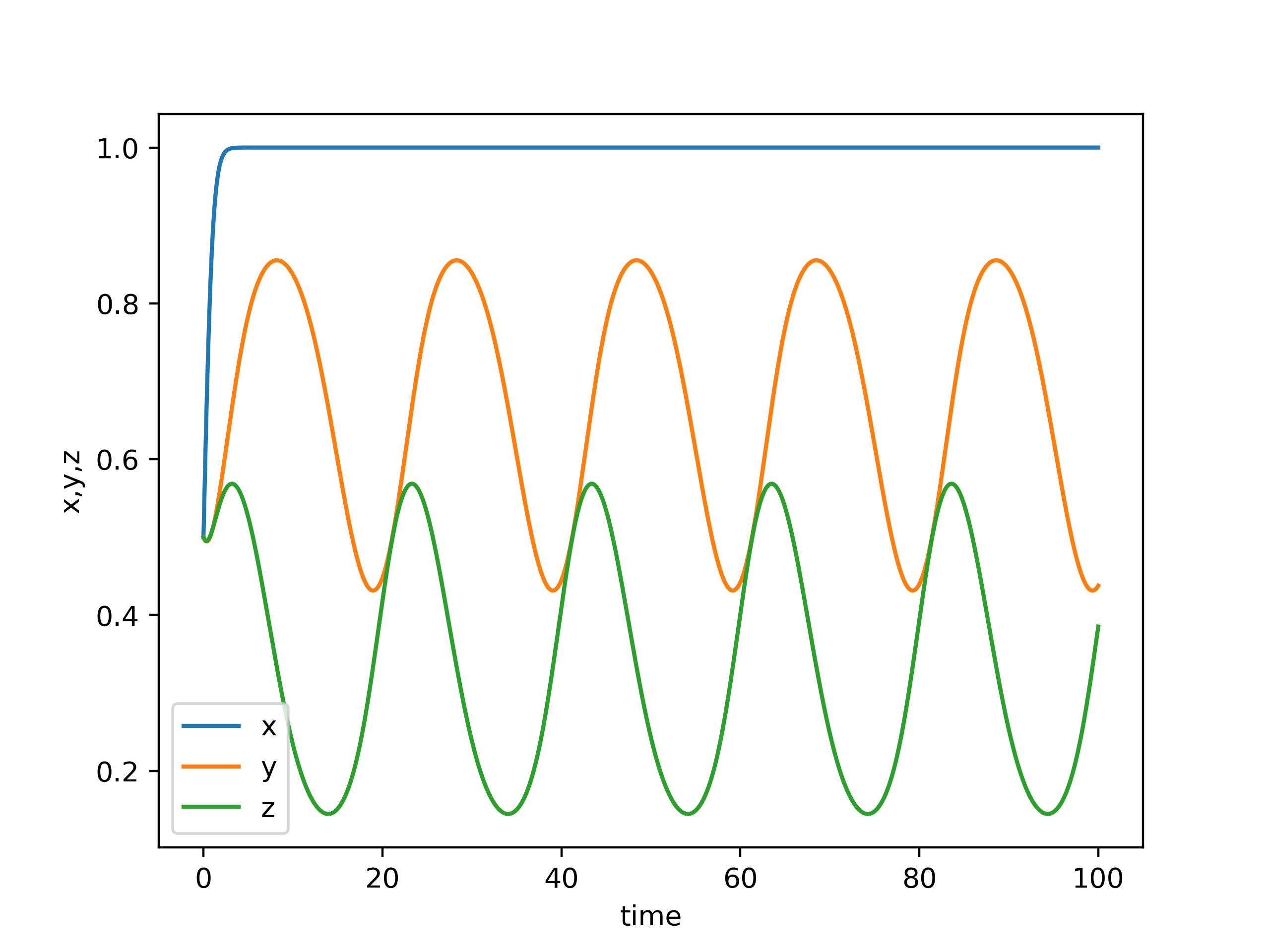}
        \caption{$x_0=y_0=z_0=0.5$, $\epsilon = 0.01$}
        \label{fig:sub1b}
    \end{subfigure}
    
    \begin{subfigure}[b]{0.4\textwidth}
        \centering
        \includegraphics[width=\textwidth]{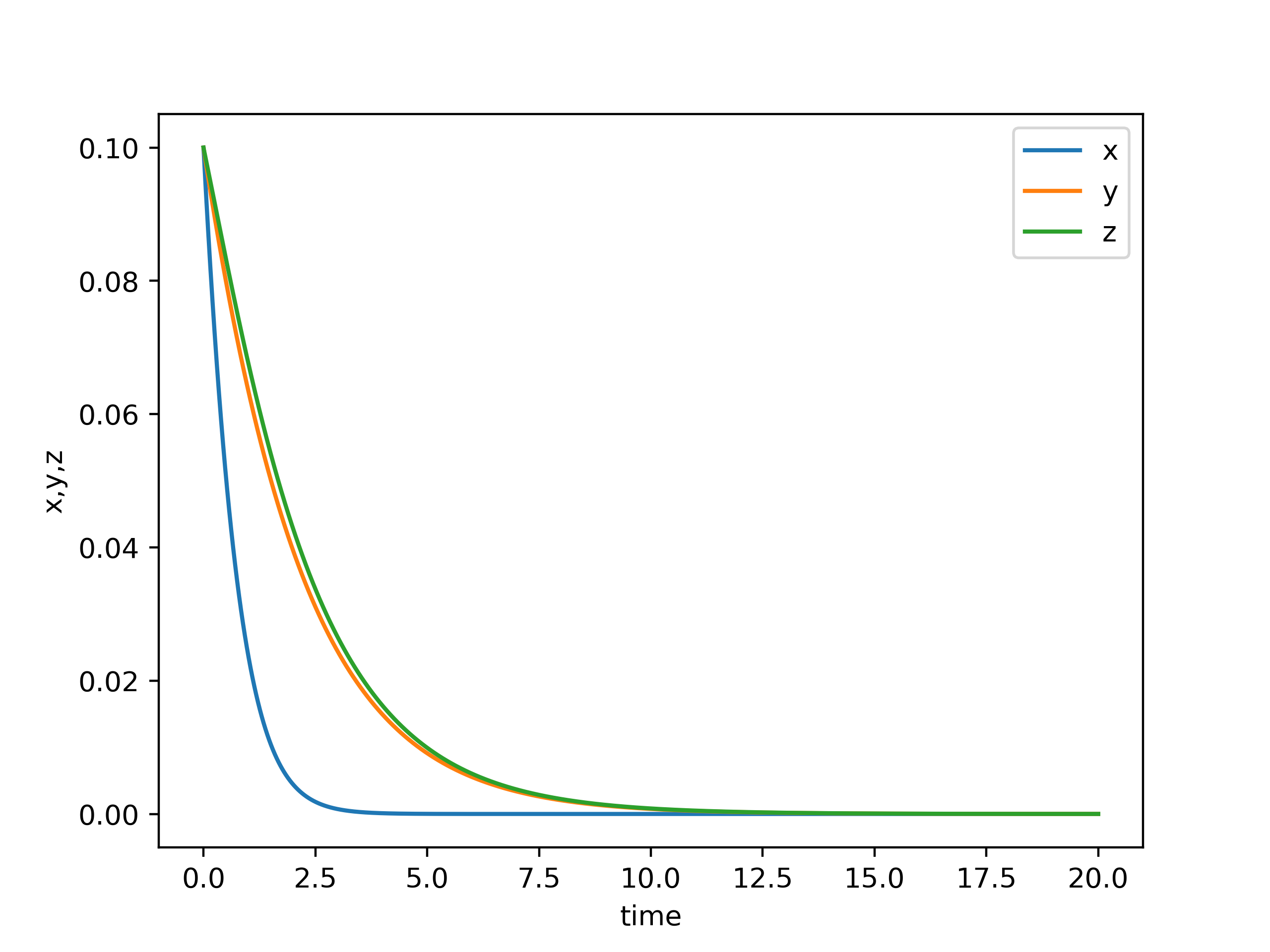}
        \caption{$x_0=y_0=z_0=0.1$,  $\epsilon = -0.5$}
        \label{fig:sub2a}
    \end{subfigure}
    \hspace{-0.3 cm}
    \begin{subfigure}[b]{0.4\textwidth}
        \centering
        \includegraphics[width=\textwidth]{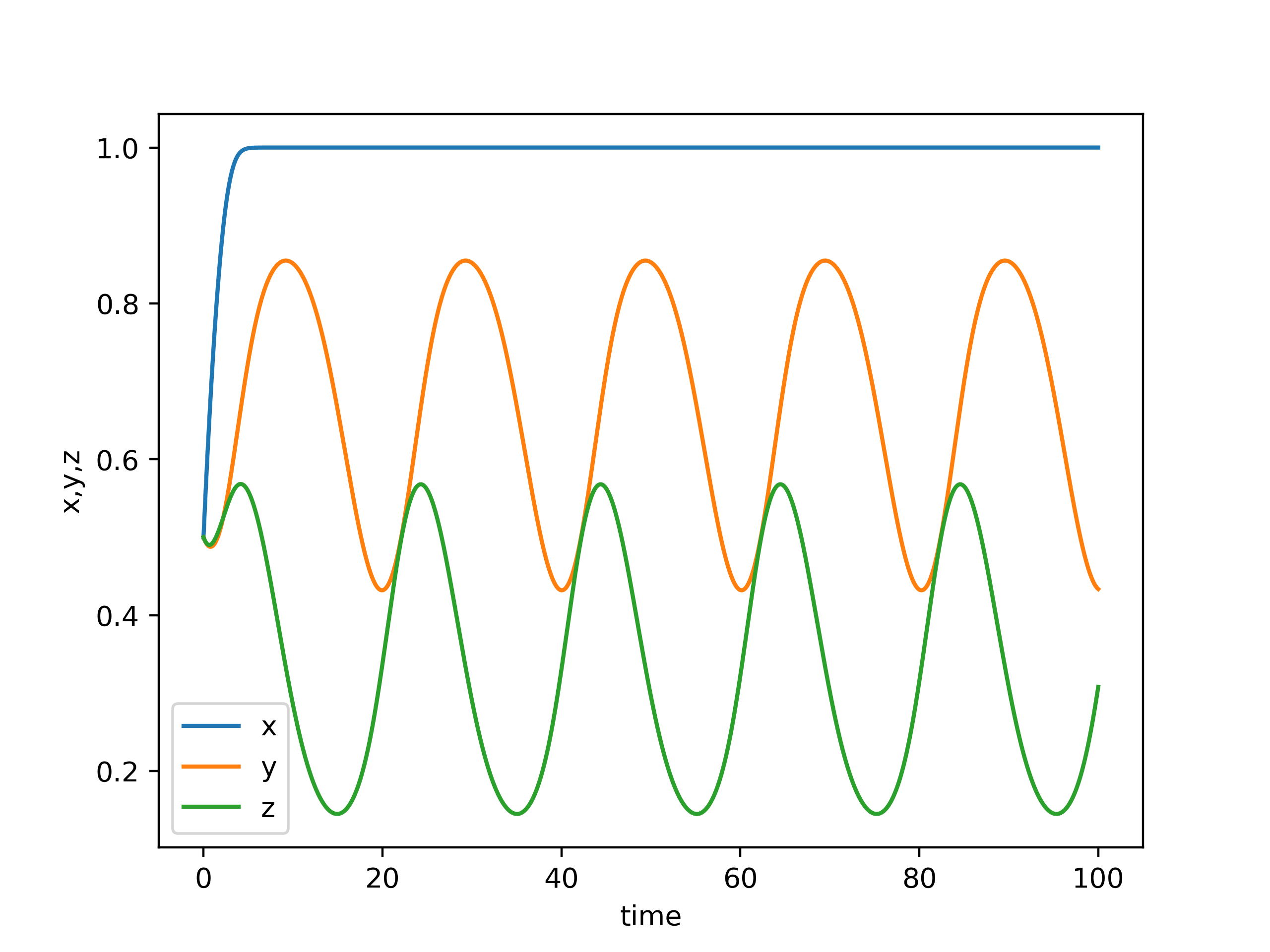}
        \caption{$x_0=y_0=z_0=0.5$,  $\epsilon = -0.5$}
        \label{fig:sub2b}
    \end{subfigure}
    
    \begin{subfigure}[b]{0.4\textwidth}
        \centering
        \includegraphics[width=\textwidth]{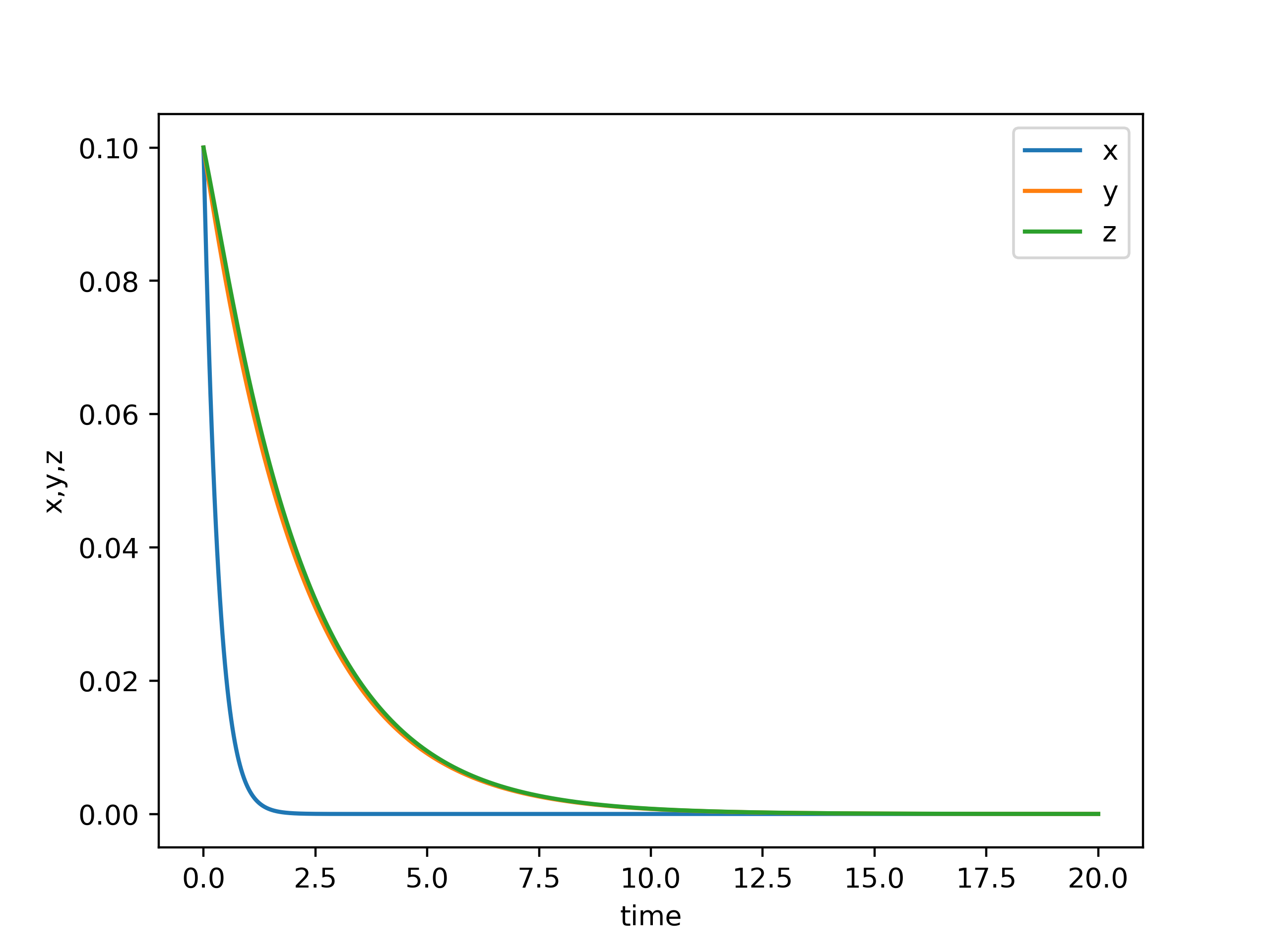}
        \caption{$x_0=y_0=z_0=0.1$,  $\epsilon = -1$}
        \label{fig:sub3a}
    \end{subfigure}
    \hspace{-0.3 cm}
    \begin{subfigure}[b]{0.4\textwidth}
        \centering
        \includegraphics[width=\textwidth]{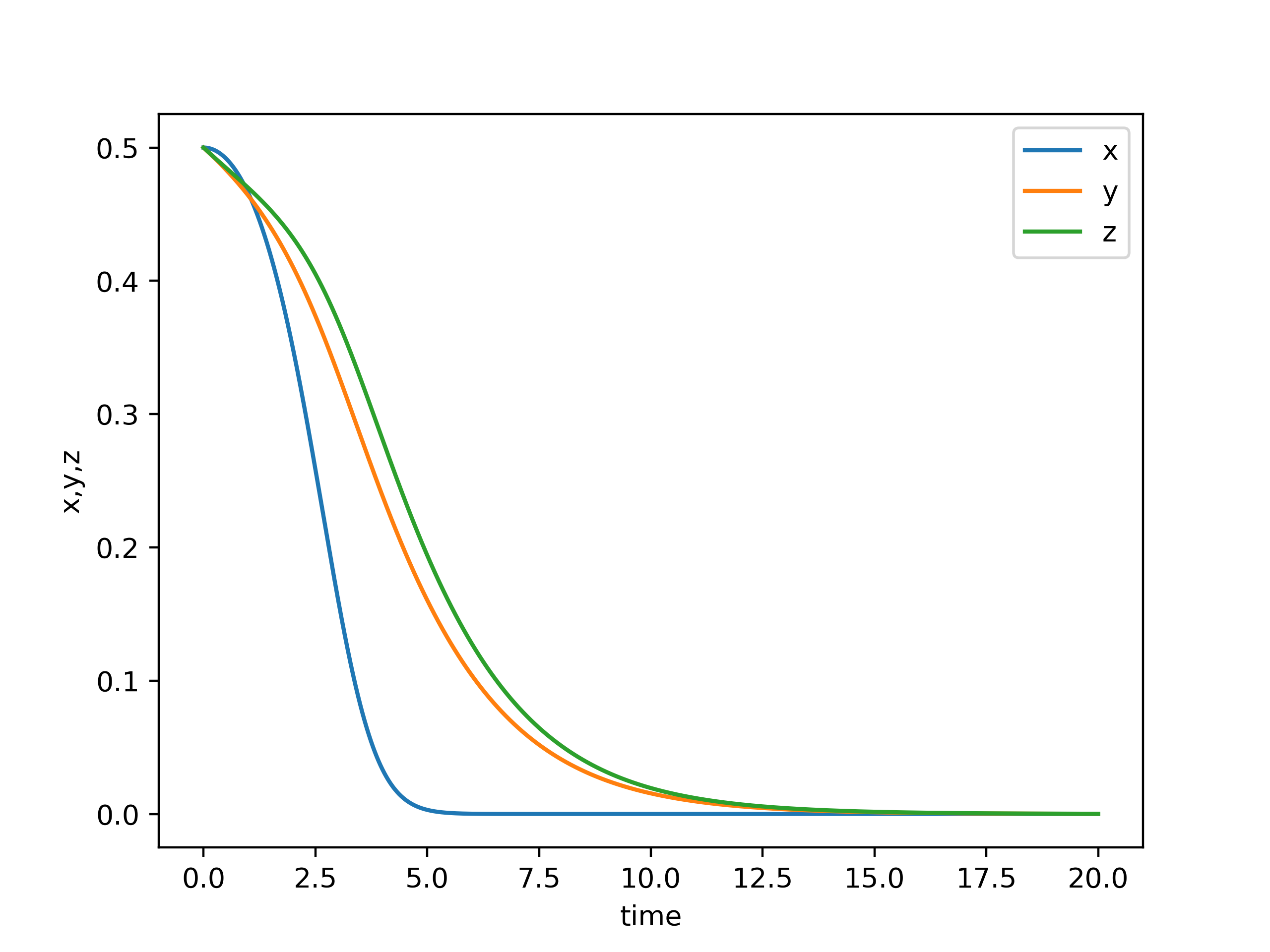}
        \caption{$x_0=y_0=z_0=0.5$,  $\epsilon = -1$}
        \label{fig:sub3b}
    \end{subfigure}
    \caption{\textbf{Numerical integration of the evolution equation for the extended model} \eqref{eq: replicator dynamics extended model}, describing the evolution of the density of trusting users $x(t)$, cooperating creators $y(t)$ and cooperating regulators $z(t)$. In all these simulations: $b_U=4$, $c_P=0.5$, $u=1.5$, $c_R=0.5$, $b_{fo} - v=1.5$. For the initial conditions and the other parameters, see the captions.}
    \label{fig_InfinitePopulation:all_images}
\end{figure}

For the extended model, using $f_C^R-f_D^R$ calculated in \eqref{eq: difference payoff R extended} in \eqref{eq: replicator dynamics baseline model},  the replicator dynamics become 
\begin{subequations}
\label{eq: replicator dynamics extended model}
\begin{align}
    \dot{x}&=x(1-x)(f^U_T-f^U_N)=x(1-x)b_U(y+\varepsilon(1-y)),\quad &x(0)=x_0\\
    \dot{y}&=y(1-y)(f^C_C-f^C_D)=y(1-y)(-c_P+u x z),\quad &y(0)=y_0\\
    \dot{z}&=z(1-z)(f^R_C-f^R_D)=z(1-z)(-c_R+x(1-y)(b_{fo}-v)),\quad &z(0)=z_0
\end{align}
\end{subequations}
where $(x_0,y_0,z_0)\in [0,1]^3$ is the initial data. Practically, compared to the baseline model, the only difference is that we allowed $v$ to be negative. This is again a complex nonlinear coupled system of $3$ ordinary differential equations.  Its solutions $x(t), y(t), z(t)$ may exhibit different behaviour depending on the values of the parameters and the initial data, see Figure \ref{fig_InfinitePopulation:all_images}.

The vertex equilibria in the baseline model are obviously still equilibria for this extended model. In addition, due to the presence of the incentive $b_{fo}$ there may be another internal equilibrium
\begin{equation}
\label{eq: internal equil extended model}
x^*=\frac{c_R(1-\varepsilon)}{b_{fo}-v},\quad  y^*=\frac{\varepsilon}{\varepsilon-1},\quad z^*=\frac{c_P(b_{fo}-v)}{u(1-\varepsilon)},
\end{equation}
if $\varepsilon<0$ and the parameters $c_P, c_R, (b_{fo}-v)$ are such that $0<x^*, y^*, z^*<1$ (in particular $b_{fo}<v$).
The Jacobian matrix is 
\begin{align*}
DF(x,y,z) =\begin{pmatrix}
b_U (1-2x)(y+\varepsilon(1-y))&b_U x(1-x)(1-\varepsilon)&0\\
u y(1-y) z& (1-2y)(-c_P+uxz)& u y(1-y) x\\
-z(1-z)(1-y)(v-b_{fo})& x(v-b_{fo}) z(1-z)& (1-2z)(-c_R-x(1-y)(v-b_{fo})).
\end{pmatrix}
\end{align*}
We are now interested in determining the stability for vertex equilibria, $(z^*,y^*,z^*)\in\{0,1\}^3$. For these equilibria, $DF(x^*,y^*,z^*)$ is always a diagonal matrix and is explicitly given by
\begin{equation*}
DF(x^*,y^*,z^*)=\begin{pmatrix}
b_U (1-2x^*)(y^*+\varepsilon(1-y^*))& 0 &0\\
0& (1-2y^*)(-c_P+ux^*z^*)& 0\\
0& 0& (1-2z^*)(-c_R-x^*(1-y^*)(v-b_{fo})).
\end{pmatrix}
\end{equation*}
$DF(x^*,y^*,z^*)$ still has three real eigenvalues, corresponding to the diagonal entries:
\[
\lambda_1=b_U (1-2x^*)(y^*+\varepsilon(1-y^*)), \quad \lambda_2=(1-2y^*)(-c_P+ux^*z^*), \quad\lambda_3=(1-2z^*)(-c_R-x^*(1-y^*)(v-b_{fo})).
\]
Thus $(x^*,y^*,z^*)$ is stable only if
\[
\lambda_1<0, \quad \lambda_2<0, \quad \lambda_3<0.
\]
If $v>b_{fo}$ then $(-c_R-x^*(1-y^*)(v-b_{fo})<0$, the stability for vertex equilibria is the same as in the case where $b_{fo}=0$.\\
We consider now $v<b_{fo}$. \\
\textbf{When $\varepsilon>0$} then $\lambda_1<0$ iff $x^*=1$, in which case $\lambda_2$ and $\lambda_3$ reduce to
\[
\lambda_2=(1-2y^*)(-c_p+uz^*), \lambda_3=(1-2z^*)(-c_R-(1-y^*)(v-b_{fo}).
\]
\begin{itemize}
    \item $(y^*,z^*)=(0,0)$. Then $\lambda_2=-c_P<0$, $\lambda_3=-c_R-(v-b_{fo})$. Thus the equilibrium $(1,0,0)$ is stable if $-c_R<v-b_{fo}<0$ and unstable otherwise.
    \item $(y^*,z^*)=(0,1)$. Then $\lambda_2=-c_P+u$, $\lambda_3=c_R+(v-b_{fo})$. Thus $(1,0,1)$ is stable only if $u<c_P$ and $c_R+(v-b_{fo})<0$ and is unstable otherwise.
    \item $(y^*,z^*)=(1,0)$. Then $\lambda_2=c_P>0$. Thus the equilibrium $(1,1,0)$ is unstable.
    \item $(y^*,z^*)=(1,1)$. Then $\lambda_2=c_P-u$, $\lambda_3=c_R>0$. Thus $(1,1,1)$ is unstable.
\end{itemize}
\textbf{When $\varepsilon<0$} 
\begin{itemize}
    \item $y^*=0$. Then $\lambda_1=b_U (1-2x^*)\varepsilon$. Hence $\lambda_1<0$ for $x^*=0$. Then $\lambda_2=-c_P<0$ and $\lambda_3=-c_R(1-2z^*)$. Hence $\lambda_3<0$ for $z^*=0$. In this case, $(0,0,0)$ is a stable equilibrium.
    \item $y^*=1$. Then $\lambda_1=b_U (1-2x^*)$. Hence $\lambda_1<0$ for $x^*=1$. Then $\lambda_2=c_P-u z^*$. Thus $\lambda_2<0$ for $z^*=0$ and $c_P<u$.  Thus $\lambda_3=-c_R<0$. If $c_P<u$ then $(1,1,0)$ is a stable equilibrium.
\end{itemize}

\subsubsection{Model with conditional trust}
For the model with conditional trust, the replicator dynamics reads
\begin{subequations}
\label{eq: replicator dynamics conditional trust model}
\begin{align}
    \dot{x}&=x(1-x)b_Uz(y+\epsilon (1-y)),\quad &x(0)=x_0\\
    \dot{y}&=y(1-y)[- c_P + uxz],\quad &y(0)=y_0\\
    \dot{z}&=z(1-z)[-c_R + b_Rx + (b_{fo}-v)x(1-y)],\quad &z(0)=z_0 
\end{align}
\end{subequations}
The stability analysis for this model is similar to the extended model (incentive for the regulator to cooperate). Hence we omit it here. 
\subsubsection{Numerical results for replicator dynamics}
We present here numerical results for the infinite population setting. 
First, Figure \ref{fig_InfinitePopulation:all_images} shows different outcomes in the evolutionary dynamics when varying initial conditions (fractions of strategists $x, y, z$ in the population). In figure \ref{fig_InfinitePopulation:all_images}a), the system first reaches trust and regulation ($TCC$), which leads to profitability for regulators to stop enforcing safety standards ($TCD$). In panels \ref{fig_InfinitePopulation:all_images}b) and d) we find a stable limit cycle between regulation and unsafe development. Users fully trust, but regulators mirror unsafe AI development. Safe AI reduces regulation, whereas unsafe AI prompts them to reinstate incentives. Thus, we could simplify the equations \eqref{eq: replicator dynamics extended model}, taking $x(t) =1$ constantly and explaining mathematically the existence of the cycle. Qualitatively, when defecting creators emerge, regulators enforce standards (because $b_{fo}>v+c_R$), thus it is profitable to comply when meeting a defecting creator. Conversely, when cooperating regulators emerge, for creators it is better to cooperate as well (as $u>c_P$); but, if creators cooperate, the enforcing standards becomes unprofitable and regulators defect. In the absence of regulation, unsafe developers re-emerge and the limit cycle restarts. In \ref{fig_InfinitePopulation:all_images}c), e) and f), trust is never established in the users' population, there is no incentive to regulate as there is no adoption, and creators ignore safety standards - $NDD$ is reached. 

\begin{figure*}
\includegraphics[scale=0.5]{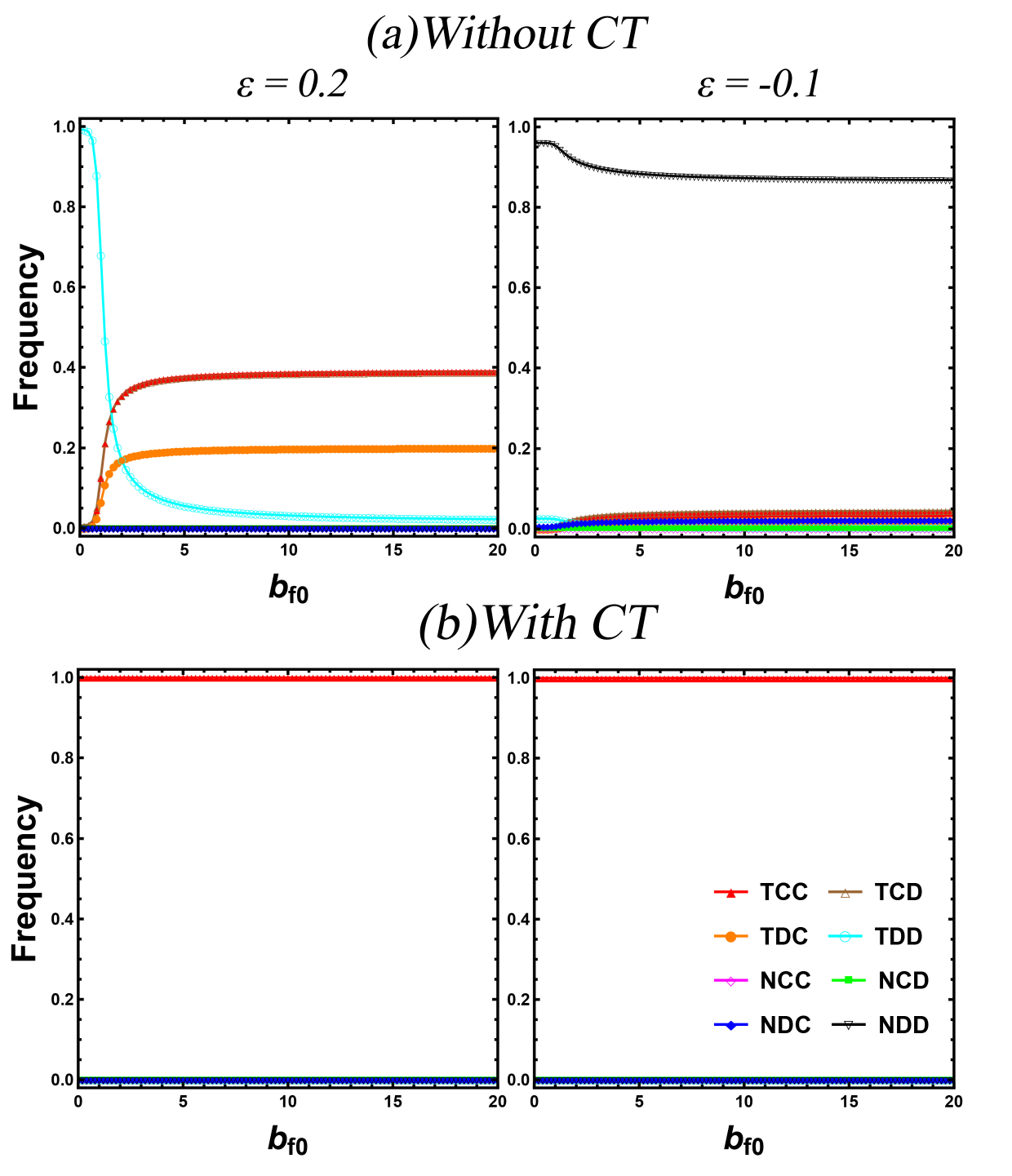}
\caption{\textbf{Low regulation cost ($c_R = 0.5$)}. Conditional trust can lead to full trust, cooperative regulation and safe development. Parameters set to: $b_U=b_R=b_P = 4$, $u=1.5$, $v=0.5$, $c_P=0.5$, $\beta=0.1$, $N_U=N_P=N_R=100$.}
\label{fig_FinitePopulation:Low_cR}
\end{figure*}

\section{Stochastic analysis}
We present here numerical results for the finite population setting (see Methods). 
Figure~\ref{fig_FinitePopulation:Low_cR} shows results for varying $b_{fo}$ for both proposed models when the cost of regulation $c_{R}$ is sufficiently small. We identify two regimes, based on the sign of the risk factor $\epsilon$. We assume that defection in the creator population (i.e., strategies of the form $xDz$) is the most deleterious outcome for users, but there still exists a potential benefit when the sign of $\epsilon$ is positive. In such cases, we observe the evolution of trust regardless of the success of regulators. Given low incentives for capturing unsafe developers ($b  _{fo}$), regulators fail to adequately enforce safety. Increasing incentives compels regulators to act, thus giving rise to a mixed outcome between $TCC, TCD$ and $TDC$ (refer to Figure \ref{fig_SD} for an in-depth view of the dominance cycles that govern these states). Whether the population evolves to $TDD$ (for low $b  _{fo}$) or the mixed state, we note that for this regime, it is always beneficial to adopt even unsafe AI (see discussion). If $\epsilon < 0$, no trust can evolve, and technology is never adopted. Importantly, When regulation is cheaply implementable (low $c_R$), conditional trust solves these issues for all incentives and in both risk regimes. 

\begin{figure*}
\includegraphics[scale=0.5]{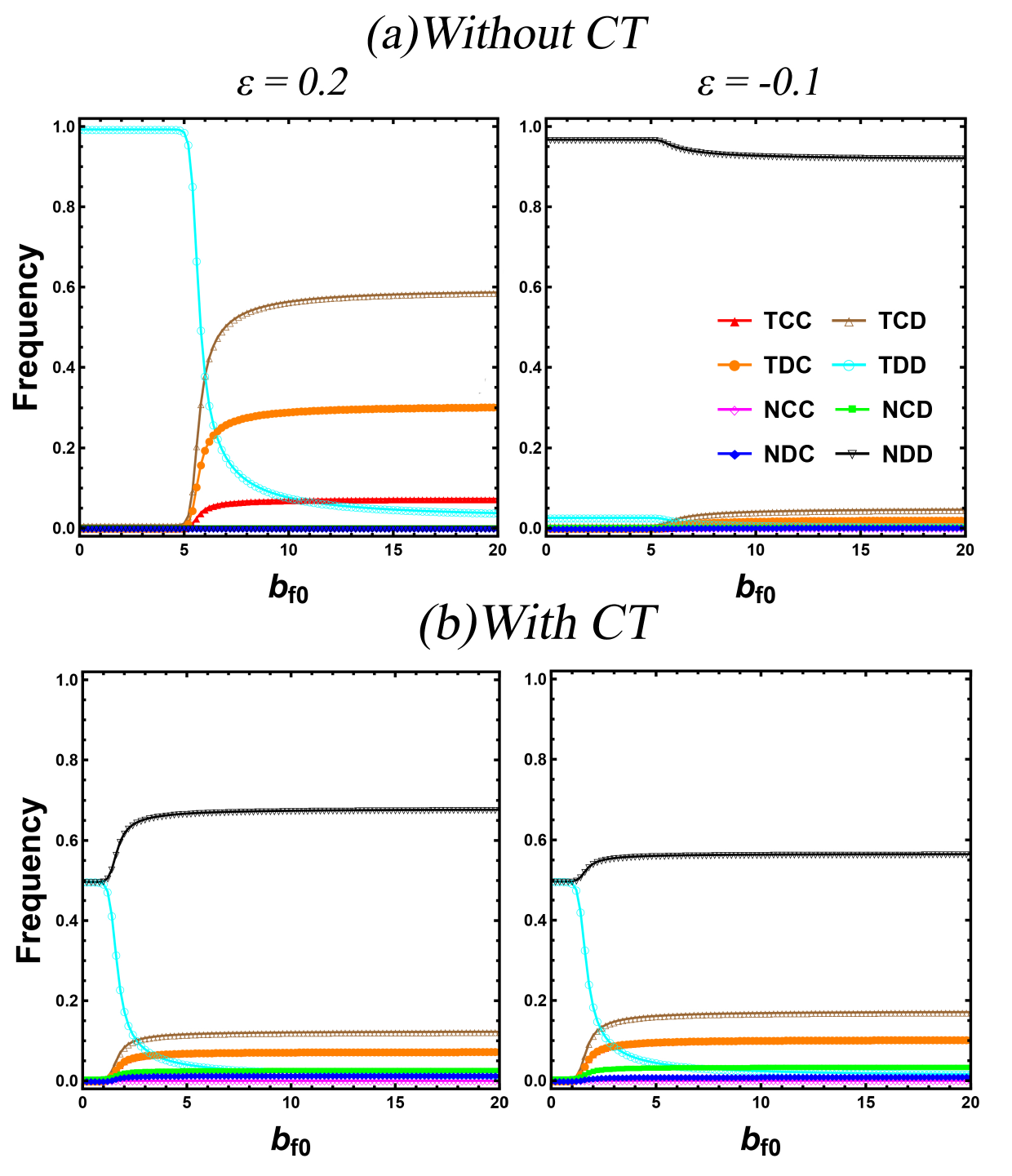}
\caption{\textbf{High regulation cost ($c_R = 5$).} Conditional trust is not sufficient to enable safe development and AI adoption when the cost of regulation is high, as this discourages regulators to cooperate. Parameters set to: $b_U=b_R=b_P = 4$, $u=1.5$, $v=0.5$, $c_P = 0.5$, $\beta=0.1$, $N_U=N_P= N_R = 100$}
\label{fig_FinitePopulation:Large_cR}
\end{figure*}

\begin{figure*}
\includegraphics[scale=0.7]{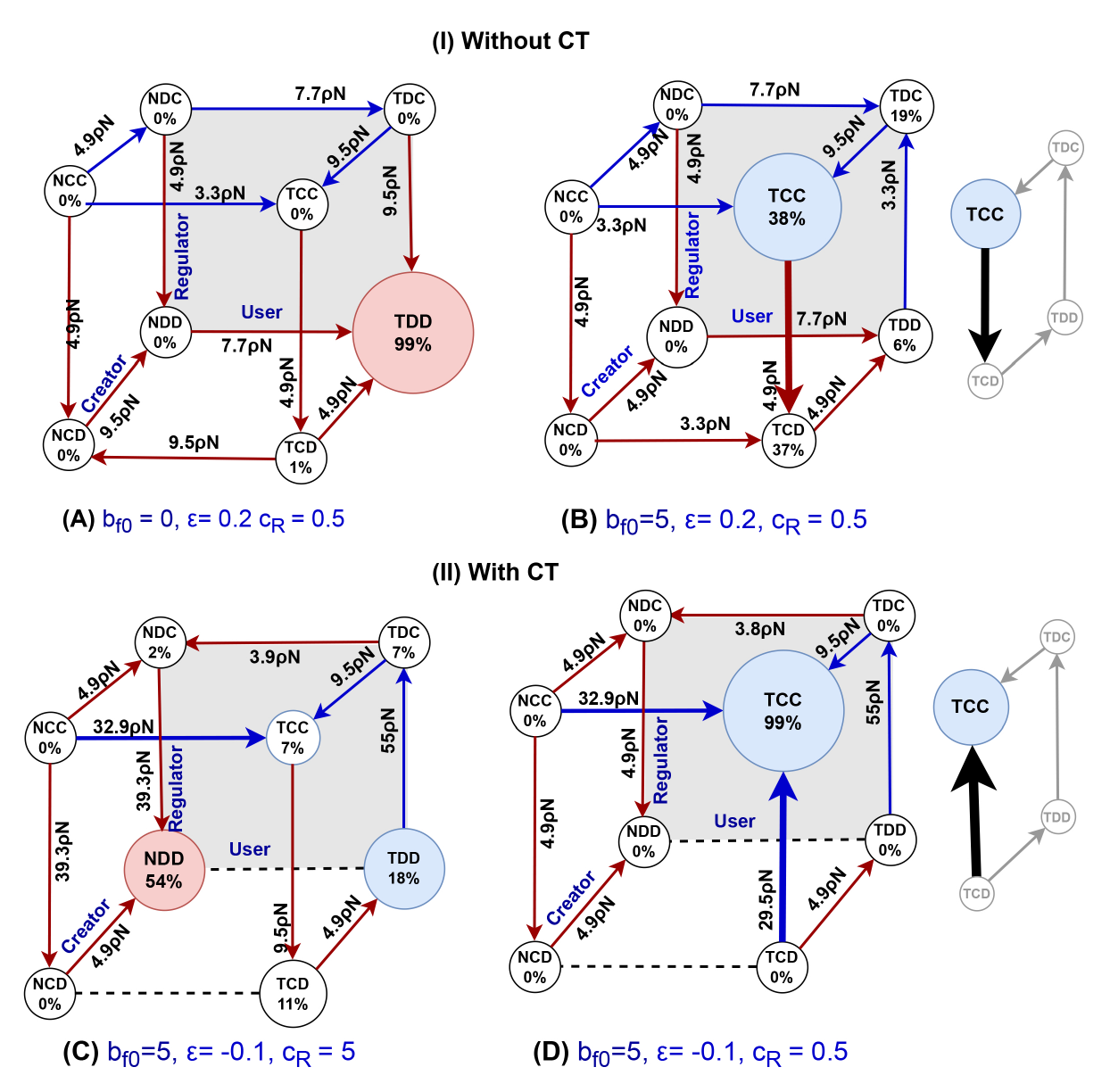}
\caption{\textbf{Stationary distribution and transitions amongst the strategies (with conditional trust and without conditional trust) ($b_R = 4$).} Only arrows where the transition probability is larger than the opposite one, are shown. Dashed lines represent neutral transitions.  Parameters set to: $b_U = b_P = 4$, $u=1.5$, $v=0.5$, $c_P = 0.5$, $\beta=0.1$. Population sizes: $N_U$ = $N_P$ = $N_R = 100$.}
\label{fig_SD}
\end{figure*}

Figure \ref{fig_FinitePopulation:Large_cR} shows results for varying $b_{fo}$ for different models, when the cost of regulation $c_{R}$ is large compared to the benefit, e.g., when the technology is highly sophisticated and it is difficult to develop technology to capture unsafe behaviour. In the regime of low risk  $\epsilon < 0$, we observe defection because users do not trust the technology and the regulators. For very low values of $b_{fo}$, there is no incentive to deviate from defection, and this remains the main regime even increasing the $b_{fo}$ values. With conditional trust, illustrated in Fig. 6 (b), even in the absence of trust in regulations, other strategies start emerging, due to cyclic dominance between $TCC \rightarrow TCD \rightarrow TDD \rightarrow TDC$ (see Figure \ref{fig_SD}).  This cycle only emerges if regulators are given a high enough incentive to enforce safety standards ($b_{fo}$). With a high risk factor $\epsilon < 0$, trust, regulation and safety never emerge. Interestingly, extending this model to conditional trust, we see a decrease in trust and adoption. Previously, low risk ($\epsilon > 0$) provided enough of a benefit for users to adopt AI even in the case of rapid, risky development, yet regulators prevent this from occurring. Regulation is too costly, thus regulators restrict access to AI, deeming it unsafe. If we consider conditional trust in such cases, users cannot elect to adopt technology unless regulators approve of creators, which hinges on the cost of regulation.  


\section{Discussion}

\subsection{Key Takewaways}
Our conceptual model of an AI governance ecosystem highlights several key factors relevant to both current and future regulatory landscapes for AI:

\begin{itemize}
\item[I.] AI companies are motivated to implement safety protocols only when regulatory bodies have the capacity to enforce compliance,
\item[II.] In scenarios where the potential negative consequences of unsafe AI development are significant, user trust in the system hinges on the dependability of regulatory authorities,
\item[III.] For regulatory bodies to consistently enforce safety measures, the cost associated with regulation must be manageable, and their capacity to identify non-compliant AI creators should be high,
\item[IV.] Critical users, namely those whose trust in the regulatory system is conditional on its perceived effectiveness, are essential to enable a stable and long-lasting trust in the system, where creators comply with regulations and regulators continue to enforce them. 
\end{itemize}

We found that trust can only emerge when regulators have an incentive to regulate. Moreover, the presence of critical users, that is, users that place their trust conditionally, are able to break the cyclic dynamics that we observe in Fig.~\ref{fig_InfinitePopulation:all_images}a. Figs.~\ref{fig_SD}b and d show how the presence of these users who trust conditionally breaks the cycle in the dimension of the regulator and creator populations, by making TCC dominant over TCD. That is, when users trust conditionally, enforcing regulation risk dominates not enforcing it.

\subsection{Implications for AI Governance}

We focus on the policy implications of our main findings III and IV: high costs in providing effective regulation pose an obstacle to building trust in future AI systems, and that it is vital to help users discern which systems have undergone effective regulation.

Scenarios where regulators face large costs in assessing the risks of AI systems are plausible. Existing AI evaluators already experience relatively high costs in providing effective regulation for fronteir AI systems. Foundation models like Open AI's GPT-4  require large amounts of computational hardware and access to the top AI talent \cite{cottier2024who}. Evaluations of these systems are likewise expensive and demand teams with expertise in AI. Academia and public institutions who wish to assess the risks of new AI systems therefore compete with AI creators in acquiring these resources. 

To overcome these barriers to entry, policy makers should consider (i) investing in building up government capacity to assess the risks of large AI systems \cite{whittlestone2021why}, and (ii) subsidising organisations that show promise in providing services that improve the effectiveness of AI regulation. Governments may therefore wish to take heed of the UK's investment in their AI Safety Institute, \citet{k2023institute}. 

Governments and regulators can benefit from communicating information that allows users to discern which methods that regulators use are reliable (so that, in turn, they can discern which AI systems are safe and trustworthy).
Doing so requires investing in the expertise needed to follow the recommendations on how to assess the risks presented by AI systems \cite{dafoe2023ai}. This could mean conducting machine learning model evaluations or other tests as part of an audit or risk assessment of new frontier AI systems \cite{shevlane2023model, koessler2023risk}. The role of academia and the commentariat should also not be underestimated here \cite{powers2023stuff}. Academia and commentators can also play a key role in communicating to users information about the effectiveness of regulations. Our model can be extended to capture this by including academics and commentators as an additional evolving population of agents. 

In the case where the regulator is a public institution with greater powers, they may wish to monitor the size of training runs for frontier AI systems or to impose fines on companies whose AI systems cause significant economic harm \cite{shavit2023what}. The governance of computing hardware (shortened to compute) has also been noted as a point of potentially high leverage, given that large amounts of physical infrastructure are much easier to monitor, track, and limit in principle than software, data, or model weights \cite{sastry2024computing}. 

One more avenue regulators can pursue to help demonstrate to users that trustworthy AI systems will be developed and deployed safely is to encourage international coordination in the governance of AI systems, perhaps by proposing designs for international institutions to govern new AI breakthroughs or AI safety projects \cite{trager2023international, ho2023international}. World powers have expressed an awareness that AI could pose enormous risks when signing the Bletchley Declaration \cite{k2023bletchley}, highlighting that international coordination of some form is tractable and desired.

Finally, public institutions will be necessary to ensure that companies and users, particularly large enterprises that serve a wide number of additional users, can form reliable impressions of the reputation of different regulators \cite{k2023institute}. Model registration is an essential prerequisite for such a reputation system; Users need to know which regulatory evaluations match which AI systems \cite{oreilly2023ItTimeCreate}. Moreover, governments may wish those providing regulatory services to preregister a set of experiments they plan to run to test the model. Regulators could later add details about how they dealt with unforeseen findings in their experiments. This would help them to demonstrate vigilance (i.e. that they have a high likelihood of finding early signs of dangerous capabilities when they in fact exist), while ensuring that their insights can be shared with other organisations quickly and safely \cite{whittlestone2021why, bova2023both}. Keep in mind that, as our results show, this information is useful whenever we have multiple organisations evaluating frontier AI systems, including at the international level \cite{ho2023international}.

\subsection{Limitations and Areas for Future Research}

Above, we discussed how decision makers can act on the insights from our results. Now, we turn to the limitations of the present model. An important simplification we made was to limit each actor in our model to a small set of strategic choices. Future research could address these limitations by incorporating partner selection into these models between populations. In a more realistic setting, users would choose which AI system they want to use. Companies too may choose to relocate to avoid especially burdensome regulation.

We also think it is essential to model more explicitly competition between different regulatory blocs. One appropriate formalism for achieving this would be to model each regulatory bloc as an island network that is connected loosely to other islands. These islands of regulation would be in competition with each other through a process of cultural group selection \cite{binmore2005natural,van2009group,richerson2016cultural}. For example, creators may move to regulatory regimes more favourable to themselves, while regulators may imitate the regulatory regimes of other groups that are more successful than themselves. 

Our model also does not fully capture the race dynamics between AI companies. OpenAI’s initial commercial breakthrough with ChatGPT has sparked interest in building more capable foundation models among the world's leading tech companies \cite{cottier2024who}. As predicted by other game-theoretic models of tech race dynamics, remaining at the frontier of AI capabilities appears to take precedence over pursuing AI safety research agendas \cite{armstrong2016racing, han2020regulate}. It is noteworthy that to date, companies have had relatively limited success making generative AI systems fundamentally safer. 

Scaling AI systems quickly encounters a brisk trade-off with safety research. It allows less time to investigate and understand the impacts of smaller models, particularly when new algorithms are used to pretrain or fine tune those models. More talent and computational resources must also be allocated to these large experiments, allowing fewer resources to pursue research agendas on AI safety. Therefore, we cannot expect even relatively safety-minded AI labs to choose the optimal trade-off, given the huge competitive dynamics at play.  For a further discussion of what can be done to directly kerb race dynamics, consult the following references \cite{shavit2023what, askell2019role, han2022voluntary, bova2023both, jensen2023industrial1}.

While we recognise the limitations of our simpler scenario, we find our approach useful as a tool for thinking through what assumptions policymakers must make to assume that different regulatory regimes are likely to achieve their policy objectives. 

Further avenues of research should test the predictions of our model, providing real-world evidence of strategic interactions between users, creators, and regulators in the AI domain. Exploring additional incentive structures for regulators could further enhance our understanding of effective AI governance. A useful incentive structure to consider is that AI companies could be mandated to pay for regulatory services that governments consider essential to demonstrate the safety of the model \cite{clark2019regulatory, hadfield2023regulatory}.

Our findings offer valuable insights into the dynamics at play in the AI regulation landscape and highlight the indispensable role of reliable regulators, incentivised by either governmental rewards or the maintenance of a prestigious reputation. They highlight the value of taking a game-theoretic approach to formally model the effects of different kinds of regulatory system on the behaviour of both AI creators and users.


\section{Appendix}



\begin{figure}[h!]
    \centering
    \begin{subfigure}[b]{0.4\textwidth}
        \centering
        \includegraphics[width=\textwidth]{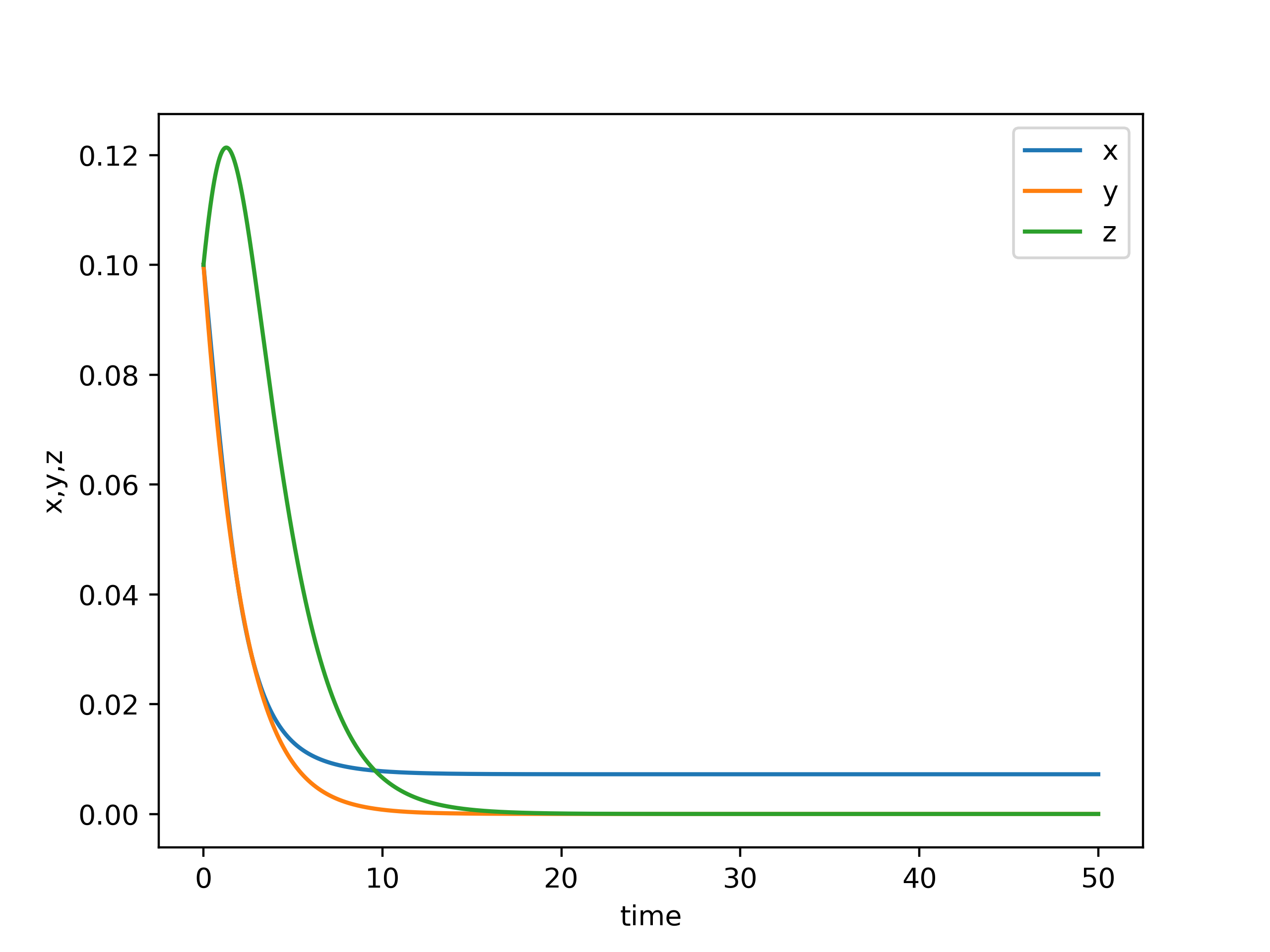}
        \caption{$x_0=y_0=z_0=0.1$, $\epsilon = -1.2$, $b_U=b_R=4$ }
        \label{fig:sub1si}
    \end{subfigure}
    \hspace{-0.3 cm}
    \begin{subfigure}[b]{0.4\textwidth}
        \centering
        \includegraphics[width=\textwidth]{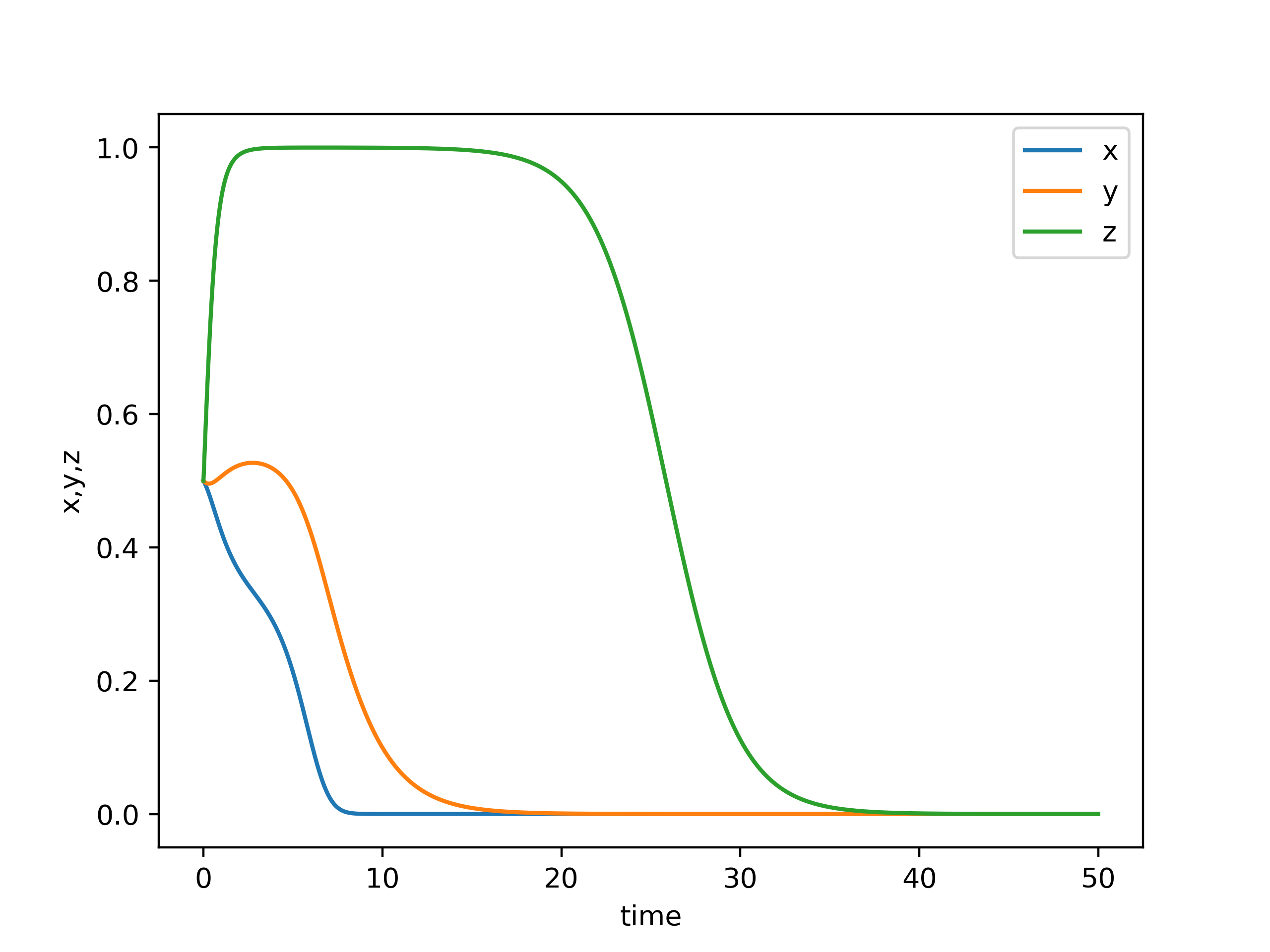}
        \caption{$x_0=y_0=z_0=0.5$, $\epsilon = -1.2$, $b_U=b_R=4$}
        \label{fig:sub2si}
    \end{subfigure}
    
    \begin{subfigure}[b]{0.4\textwidth}
        \centering
        \includegraphics[width=\textwidth]{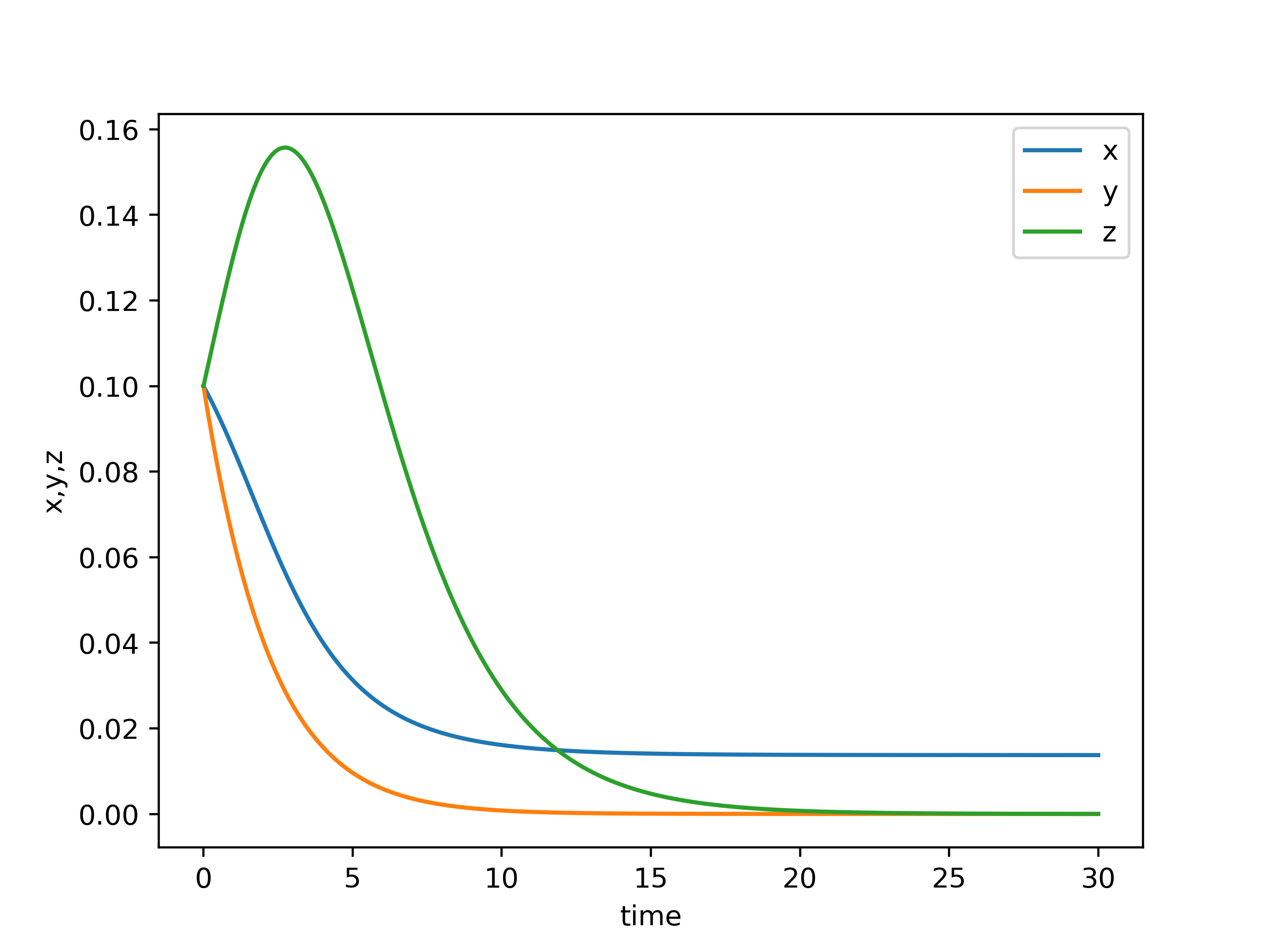}
        \caption{$x_0=y_0=z_0=0.1$, $\epsilon = -0.5$, $b_U=b_R=4$}
        \label{fig:sub3si}
    \end{subfigure}
    \hspace{-0.3 cm}
    \begin{subfigure}[b]{0.4\textwidth}
        \centering
        \includegraphics[width=\textwidth]{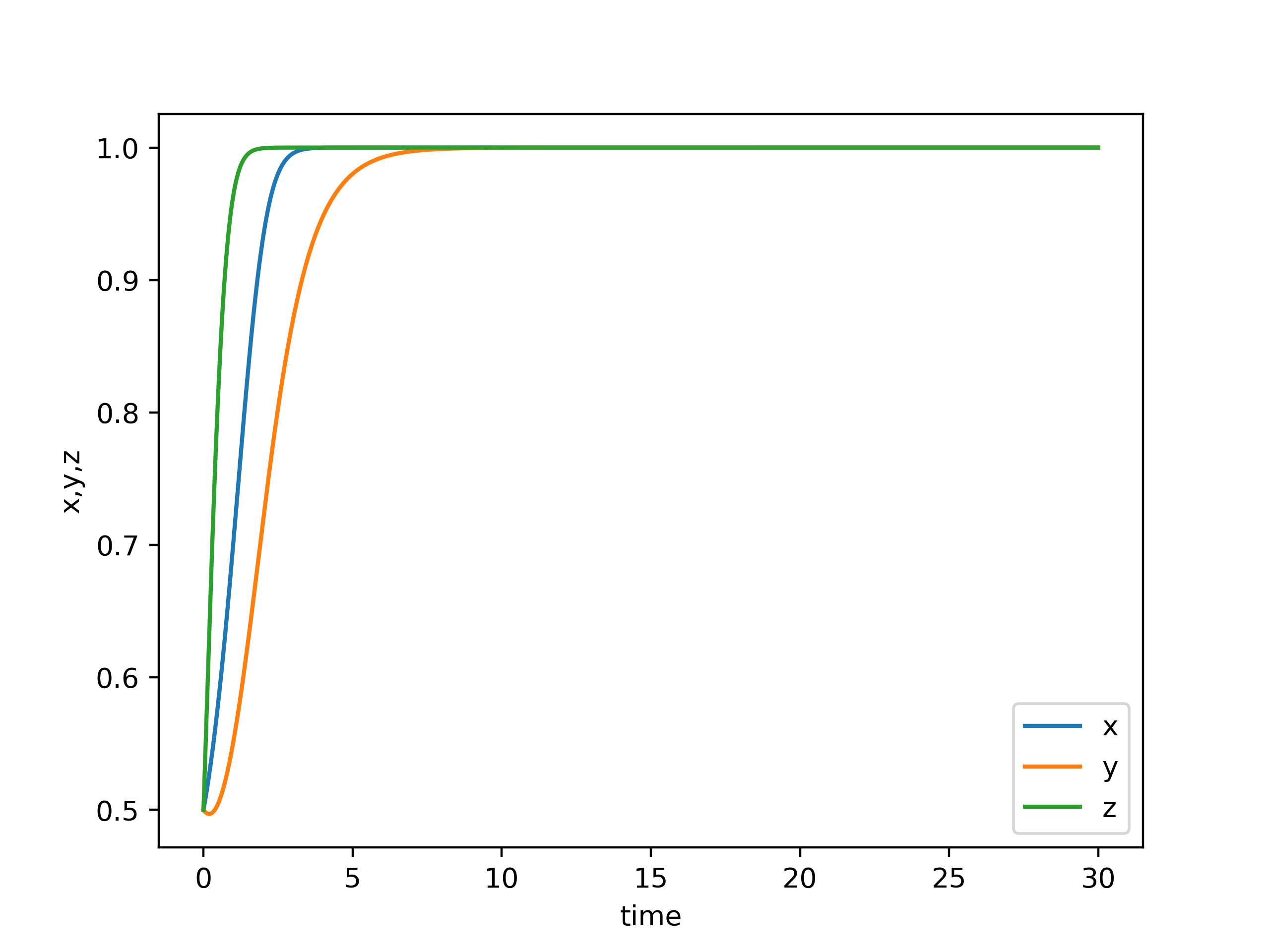}
        \caption{$x_0=y_0=z_0=0.5$, $\epsilon = -0.5$, $b_U=b_R=4$}
        \label{fig:sub4si}
    \end{subfigure}
    
    \begin{subfigure}[b]{0.4\textwidth}
        \centering
        \includegraphics[width=\textwidth]{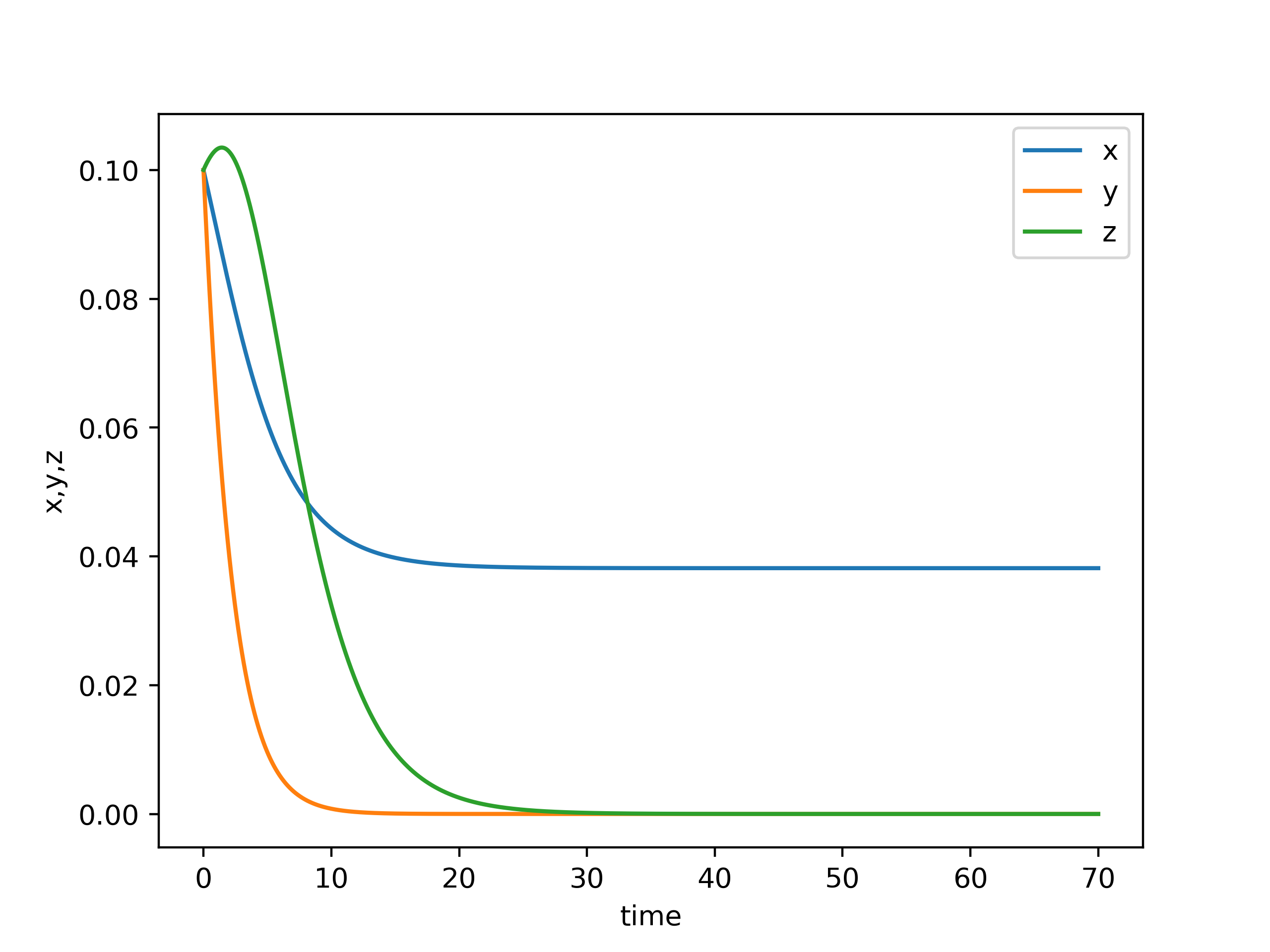}
        \caption{$x_0=y_0=z_0=0.1$, $\epsilon = -1.2$, $b_U=b_R=1$}
        \label{fig:sub3si}
    \end{subfigure}
    \hspace{-0.3 cm}
    \begin{subfigure}[b]{0.4\textwidth}
        \centering
        \includegraphics[width=\textwidth]{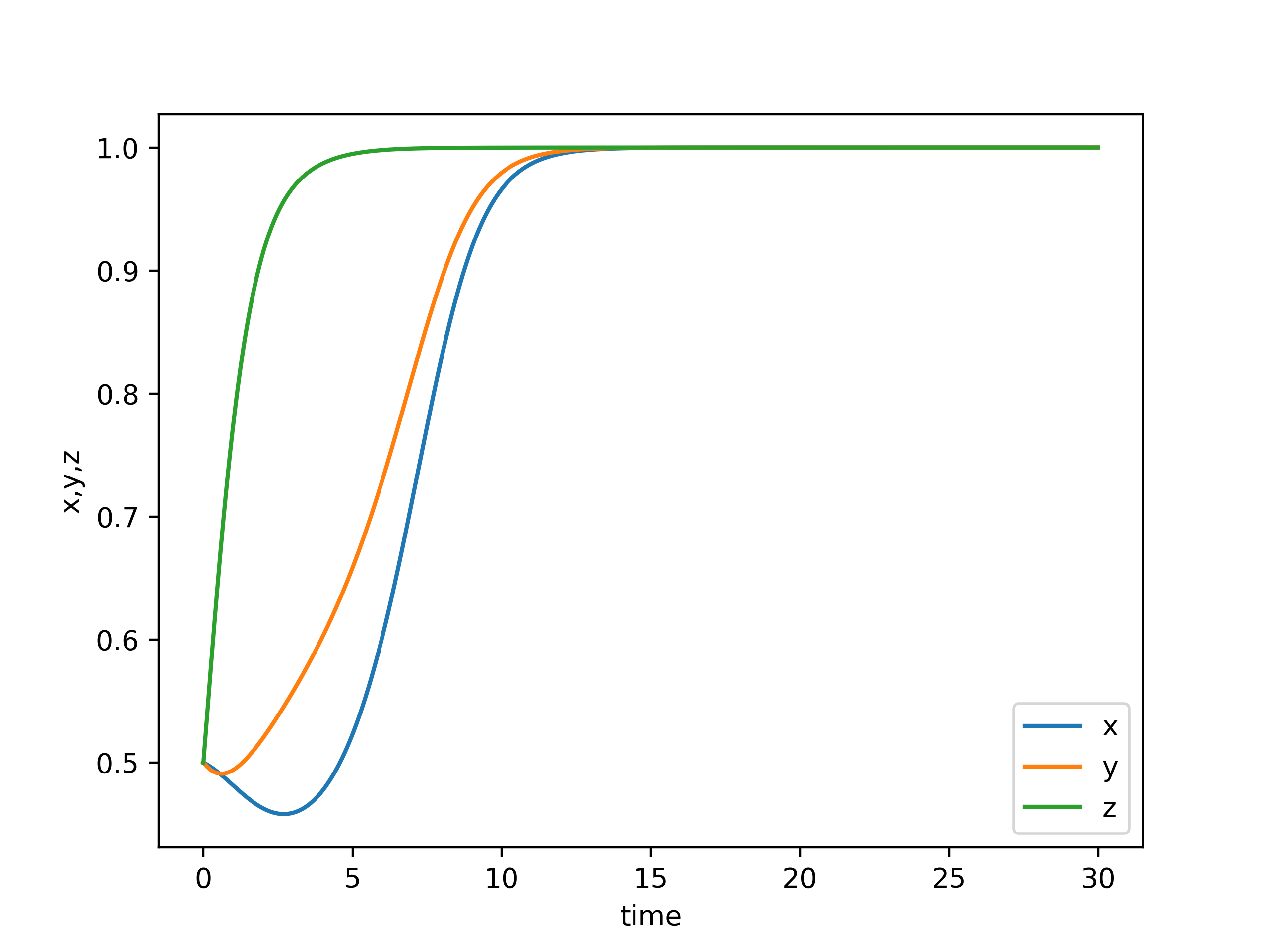}
        \caption{$x_0=y_0=z_0=0.5$, $\epsilon = -1.2$, $b_U=b_R=1$}
        \label{fig:sub4si}
    \end{subfigure}
    \caption{Numerical integration of the evolution equation for the model with conditional trust \eqref{eq: replicator dynamics conditional trust model}, describing the evolution of the density of trusting users $x(t)$, cooperating creators $y(t)$ and cooperating regulators $z(t)$. In all these simulations: $c_P=0.5$, $u=1.5$, $c_R=0.5$, $b_{fo} - v=5$. For the initial conditions and the other parameters, see the captions.}
    \label{fig_InfinitePopulation:all_images_appendixCT}
\end{figure}

\section*{Acknowledgement}
This work is produced during the workshop "AI Governance Modelling", funded through the generous support from the Future of Life institute (TAH). 
\bibliographystyle{IEEEtran}
\bibliography{refs} 



\end{document}